\documentclass[a4paper,fleqn,review]{cas-sc}

\usepackage{cleveref}
\usepackage{textcomp}
\usepackage{makecell}
\usepackage[numbers,sort&compress]{natbib}
\usepackage{bm} 

\def\tsc#1{\csdef{#1}{\textsc{\lowercase{#1}}\xspace}}
\tsc{WGM}
\tsc{QE}
\def\R{\mathbb{R}}
\def\W{\mathbf{W}} 
\def\b{\mathbf{b}} 
\def\T{\mathsf{T}} 

\newdefinition{rmk}{Remark}
\newproof{pf}{Proof}
\newproof{pot}{Proof of Theorem \ref{thm}}

\begin{document}

\let\WriteBookmarks\relax
\def\floatpagepagefraction{1}
\def\textpagefraction{.001}

\shorttitle{Physics-informed neural networks for parameter learning of wildfire spreading}

\shortauthors{Vogiatzoglou, et al.,}  

\title [mode = title]{Physics-informed neural networks for parameter learning of wildfire spreading}



%

\author[1]{K. Vogiatzoglou}
[       orcid=0009-0004-8333-247X]


\cortext[1]{Corresponding author}

\cormark[1]

\ead{kvogiatzoglou@uth.gr}


\credit{Data Curation, Formal analysis, Software, Validation, Visualization, Writing - Original Draft}

\affiliation[1]{organization={System Dynamics Laboratory, Department of Mechanical Engineering, University of Thessaly},
            city={Volos},
            postcode={38334}, 
            country={Greece}
}

\author[1]{C. Papadimitriou}


\ead{costasp@uth.gr}


\credit{Supervision, Writing - Review \& Editing}

\affiliation[2]{organization={Transport Processes \& Process Equipment Laboratory, Department of Mechanical Engineering, University of Thessaly},
            city={Volos},
            postcode={38334}, 
            country={Greece}
}

%


\author[2]{V. Bontozoglou}


\ead{bont@uth.gr}


\credit{Methodology, Writing - Review \& Editing}

\affiliation[3]{organization={Automatic Control \& Autonomous Systems Laboratory, Department of Mechanical Engineering, University of Thessaly},
            city={Volos},
            postcode={38334}, 
            country={Greece}
}

\author[3]{K. Ampountolas}




\ead{k.ampountolas@uth.gr}


\credit{Conceptualization, Methodology, Project Administration, Visualization, Writing - Review \& Editing}

\begin{abstract}
Wildland fires pose a terrifying natural hazard, underscoring the urgent need to develop data-driven and physics-informed digital twins for wildfire prevention, monitoring, intervention, and response. In this direction of research, this work introduces a physics-informed neural network (PiNN) designed to learn the unknown parameters of an interpretable wildfire spreading model. The considered modeling approach integrates fundamental physical laws articulated by key model parameters essential for capturing the complex behavior of wildfires. The proposed machine learning framework leverages the theory of artificial neural networks with the physical constraints governing wildfire dynamics, including the first principles of mass and energy conservation. Training of the PiNN for physics-informed parameter identification is realized using synthetic data on the spatiotemporal evolution of one- and two-dimensional firefronts, derived from a high-fidelity simulator, as well as empirical data (ground surface thermal images) from the Troy Fire that occurred on June 19, 2002, in California. The parameter learning results demonstrate the predictive ability of the proposed PiNN in uncovering the unknown coefficients of the wildfire model in one- and two-dimensional fire spreading scenarios as well as the Troy Fire. Additionally, this methodology exhibits robustness by identifying the same parameters even in the presence of noisy data. By integrating this PiNN approach into a comprehensive framework, the envisioned physics-informed digital twin will enhance intelligent wildfire management and risk assessment, providing a powerful tool for proactive and reactive strategies.    
\end{abstract}


\begin{highlights}
\item A physics-informed neural network (PiNN) predicts wildfire spreading.

\item A set of key model parameters captures the complex behavior of wildfires.

\item PiNNs integrate training data while adhering to wildfire system dynamics.

\item Efficacy of PiNNs in 1D/2D firefronts using synthetic and empirical fire event data.

\end{highlights}


\maketitle


\section{Introduction}
\label{sec:Introduction}

Wildland fires represent instances of aggressive fire expansion with a profound impact on shaping the characteristics of biotic ecosystems. While they foster the natural survival of various life forms through nutrient cycling, habitat diversity, and seed germination, they also lead to disproportionate implications across multiple facets of living entities in both social, economic, and environmental domains \citep{NegarElhamiKhorasani2022,ACarvalho2011} due to their extreme progression rates. The climate crisis, exacerbated by greenhouse gas effects, global warming, and water scarcity, along with detrimental human interventions, including unfavorable land use at the wildland-urban interface (WUI) and increased forest flammability, highlight wildfires as an exceedingly frequent and severe disaster, confined to a finite set of trustworthy simulation software capable of predicting their spread.

The primary focus of a holistic simulation tool should always prioritize human safety and the preservation of natural viability. As a naturally complex phenomenon, fire behavior models primarily aim to anticipate the direction and intensity of fire spread rates by blending historical (offline) data with continuously evolving geospatial and environmental (online) data sources. However, the acquisition of online data during fire propagation presents significant challenges, such as incapacity, sparsity, and costliness. Therefore, incorporating synthetic data, along with actual data from past or ongoing wildfire events, into well-designed simulation software is essential. Despite significant efforts to develop a unified digital tool to interpret the multifaceted dynamics involved \citep{Goodrick2022, EPastor2003, Finney2013}, the extreme threat posed by wildfires remains an ongoing issue within the field of natural science. Consequently, there is a growing demand for a deeper understanding of the core principles governing these natural processes \citep{FrankmanD2013}.

Fire experts recognize that in emergent situations, empirical (including statistical) models are the most operationally oriented approaches, as they provide promptly information about the rate of spread (ROS) and direction of the firefront without requiring any detailed theoretical background \citep{EPastor2003, Rothermel1972, Marino2012}. Due to their "rule-of-thumb" nature, these models instantly leverage measurements of the ongoing fire event (and/or historical fires) to provide generalized recommendations concerning coordination, evacuation, and suppression \citep{Cruz2020,Vega1998}. Even so, their applicability falls short during extreme-danger operating conditions (e.g., spotting-moving fires, propagation over undulating terrain, and steep slopes), and they fail to represent the full functional range under constant environmental variations. Conversely, and according to our envisioned modeling approach \citep{Vogiatzoglou2024}, physics-based formulations appear to compactly integrate fundamental aspects of fluid mechanics, heat transport, and reaction kinetics theory \citep{Prieto2017, Buerger2020, Mandel2008,Hanson2000} in an attempt to reflect the actual behavior of a forest fire. Their primary purpose entails the interpretation of the physics behind fire spread and the identification of combustion-driven effects and pyroconvective interactions \citep{Simeoni2011}. In this regard, the developed modeling approach employs computationally efficient methods while delivering faster-than-real-time simulations of wildfire spread with high accuracy \citep{Vogiatzoglou2024, Coen2007}. However, the fidelity of these predictions relies on examining the inherent randomness present due to modeling assumptions and data misinformation. Uncertainty quantification is essential for reducing both aleatoric (i.e., noisy data) and epistemic (i.e., scattered and limited data) uncertainties. At the other end of the spectrum, computational fluid dynamics (CFD) simulations represent the third modeling category, encompassing all spatial and temporal scales while allowing the interaction of fuel, fire, and atmosphere \citep{Linn2005,JMCanfield2014, Coen2013}. These models offer high-resolution, three-dimensional simulations (wildfire spreading solver) under dynamic weather conditions and complex terrain features, although they necessitate extended memory allocation and computing power. Furthermore, they can capture both the updraft movement of plume (atmospheric boundary-layer solver) \citep{Tory2018} and provide detailed insights into the combustion process (combustion solver) \citep{Morvan2004}.

Generally, the management of wildfires imposes advanced predictive resources within a physics-constrained and data-driven simulation software \citep{Mandel2008}. This innovative modeling approach acts as the centerpiece of a comprehensive digital tool, facilitating coordination among communities by delivering immediate responses within a hazard and risk assessment framework. However, prediction accuracy is hindered by uncertainties stemming from both the model derivation (physical and chemical abstractions) and the incorporation of noisy data. To address these challenges, it's imperative to integrate measurements of acceptable precision, acquired through remote sensing methods (e.g., satellite, aircraft, drone, and weather radar strategies \citep{Lareau2022}), into the simulation workflow. A data-informed wildfire decision support system should seamlessly combine the power of fast predictive execution with the benefits of a physics-based formulation, enabling accurate forward estimations within an efficient parameter learning approach \citep{Vogiatzoglou2024}, thus further strengthening the overall proactive and responsive tactics under real-time limitations.

Among the preceding assumptions, parameter learning emerges as a notable enhancement toward acquiring a more profound insight into this ecological challenge. Numerous literature reviews emphasize the necessity of incorporating both synthetic \citep{Allaire2021} and real data \citep{Heui2012, Alessandri2021, Joshi2021, Shadrin2024} using diverse methodologies for parameter identification \citep{Ambroz2018, Zhang2017, Zhang2019}. Empirical correlations combined with level-set methods \citep{Alessandri2021, Lautenberger2013} have been proposed for model calibration, collecting measurements from the firefront shape, wind flow characteristics, and fuel properties. These approaches minimize a least-squares objective function that quantifies the difference between simulated and observed data at multiple time instants. Ensemble Kalman filtering \citep{Mandel2008, Rochoux2014, Coen2007}, along with Monte Carlo simulations \citep{ Xue2012}, have been suggested to include uncertainties in both measurements and predictions, update vital model parameters, and deliver data-driven wildfire forward estimations. Nonetheless, these methods struggle to accommodate real noisy data and are restricted due to dimensionality constraints, highlighting the requirement for a more cohesive methodology.

Under the concept of these limitations, a novel framework of inverse optimization algorithms has emerged, known as physics-informed neural networks (PiNNs) \citep{Lai2021,Lagaris1998,Raissi2018, Karniadakis2021}, amplifying the range of conventional parameter learning methods. 
Building on this, the present work aims to apply and demonstrate the effectiveness of PiNNs specifically for learning the parameters of a wildfire spreading model \citep{Vogiatzoglou2024}. PiNNs represent data-driven machine learning approaches that merge the theory of artificial neural networks (ANNs) with the physical constraints of real-world systems, restricting the spectrum of feasible model predictions. This integration represents the first contribution in the literature to the learning process of the unknown (hard to assess) model parameters that capture essential physical and chemical properties of the firefront. The proposed methodology for physics-informed parameter identification of a wildfire spreading model includes two fundamental constituents: (i) an artificial neural network, also recognized as a universal approximator \citep{Hornik1989}, parameterized by weights and biases; and (ii) the adherence to the physical laws of the dynamical wildfire spreading model \citep{Vogiatzoglou2024}, as described by ordinary or partial differential equations (ODEs, PDEs) \cite{Lagaris1998}. Effective training of PiNNs requires robust data acquisition methods to generate an adequate training dataset, considering both the size and quality of the data points. To this end, the developed PiNN is trained using both synthetic and empirical fire event data from the Troy Fire in California, 2002. As for the modeling formulation, it is grounded in the first principles of mass and energy conservation, while data points related to temperature and fuel measurements are incorporated during the training phase. This approach enables the network to systematically explore the full range of available parameters and ultimately converge to their optimal values.

The remainder of this paper is structured as follows. \Cref{sec:Modeling} provides a concise overview of the wildfire model derivation and delineates the essential model parameters for learning. \Cref{sec:PiNNs} introduces the streamline configuration of PiNNs and details the overall training process. \Cref{sec:Application_and_Results} outlines some significant case studies, focusing on validating the effectiveness of PiNNs for both synthetic and empirical data from the Troy Fire in California. Finally, \Cref{sec:Discussion_and_Conclusions} presents the conclusions drawn and offers suggestions for future enhancements.

\emph{Notation}: The set of real numbers is denoted by $\R$. The dot product operator between two vectors $\mathbf{a} = [a_1\,\, a_2]^\T$ and $\mathbf{b} = [b_1\,\, b_2]^\T$ is represented by $\left< \cdot \right>$ and yields $\textbf{a} \cdot \textbf{b} = a_{1}b_{1} + a_{2}b_{2}$. The gradient operator, $\nabla_{xy}$, is expressed as $\nabla_{xy}  = \frac{\partial }{\partial x}\hat{i} + \frac{\partial }{\partial y}\hat{j}$, where $\hat{i}$ and $\hat{j}$ represent the unit vectors in the $x-$ and $y-$directions, respectively. Moreover, $\nabla_{xy}^2 = \frac{\partial^2 }{\partial x^2} + \frac{\partial^2}{\partial y^2}$ stands for the Laplace operator. Eventually, when $f_i$ is mentioned, it indicates the derivative of the state variable $f$ with respect to $i$. Similarly, $f_{ii}$ denotes the second derivative of $f$ with respect to $i$. State variables for the considered wildfire model include the temperature ($T$), the endothermic fuel ($E$), and the exothermic fuel ($X$). 

\section{Wildfire spreading modeling}
\label{sec:Modeling}

This section briefly presents our interpretable physics-based wildfire model \cite{Vogiatzoglou2024} to emulate wildfire expansion and its crucial model parameters for physics-informed parameter identification. 

\subsection{Model formulation}
\label{sec:Model_Formulation}

Wildfire modeling necessitates the incorporation of diverse spatiotemporal scales and natural processes, alongside the intricate chemistry of burning fuel and the detailed physics governing fluid flow and heat transport \citep{Sullivan2009, Goodrick2022}. In this streamlined modeling configuration, the reaction kinetics include first-order Arrhenius expressions for water dehydration and wood combustion. Heat transfer is modeled by a combination of both radiation, convection, and dispersion (diffusion) \citep{Simeoni2011, Finney2013}. These mechanisms attempt to describe the complex physical characteristics induced by the interaction of airflow fluctuations above and within the plantation and buoyant flame instabilities \citep{Finnigan2007}. Moreover, specific topographical features may result in zones dominated by chaotic turbulent vortices (turbulent diffusion) generated by flame buoyant forces \citep{Finney2010, Finney2015}, which strongly affect the direction and intensity of the ongoing wildfire event by formulating updraft (bursts) and downdraft (sweeps) flame movements.

The burning fuel is considered a solid material composed of two elements: water and combustibles \citep{FranciscoJSeron2005, Morvan2004}. Specifically, the mass fraction of water, denoted as $E$, encompasses the internal content of the plant (live humidity) and the water condensed on the plant (dead humidity). The remaining mass fraction of combustibles, denoted as $X$, comprises both volatiles (e.g., CO, CO$_2$, NO$_x$, and VOC) and charcoal. Following a fundamental framework, combustion initiates with water evaporation (as the endothermic phase), progresses through volatilization and charring reactions, and ultimately terminates with oxidation reactions (as the ensuing exothermic phase) \citep{Sullivan2017, RubenSudhakarDhanarathinam2011}. Thus, the endothermic (gaseous) part of the combustion process is modeled using the Arrhenius kinetics formulation as:
\begin{equation}
\label{eq:Arrhenius1}
r_{\rm e} =c_{{1}} \, e^{-\frac{b_{1}}{T}}.
\end{equation}
The same equation is applied to the exothermic (solid) phase of the combustion process. However, the possible shortage of oxygen modifies the overall exothermic reaction term \citep{Leckner1999}, leading to:
\begin{equation}
\label{eq:Arrhenius2}
r_{\rm x} = \frac{c_{2} \, e^{-\frac{b_2}{T}}r_{\rm o}}{c_{2} \, e^{-\frac{b_2}{T}}+r_{\rm o}},
\end{equation}
where $T$ ([=] K) represents the temperature of the fire layer, $c_{1}, c_{2}$ ([=] s$^{-1}$), along with $b_1, b_2$ ([=] K), signify crucial parameters of the ignition and combustion process. Additionally, $r_{\rm o}$ ([=] s$^{-1}$) embodies the presence or absence of oxygen (frequency of incoming oxygen) during the pyrolysis stage.

The proposed model implements a novel method to separate the thermal energy from the fluid mechanics aspect (momentum conservation) and facilitates simplified versions of combustion kinetics by focusing on key reaction pathways and mass-energy balances to deliver insights into temperature dynamics and fuel composition effects. It employs a computationally inexpensive system of differential equations, which empowers simple calculations while maintaining acceptable accuracy \citep{Mandel2008, Buerger2020, Hanson2000, Simeoni2011}. Additionally, this straightforward approach incorporates fundamental physics articulated by a finite set of key model parameters, making it adaptable to sensory input data \citep{Lareau2022, Nicholas_McCarthy2019}, as well as enabling data-informed estimations. These advancements contribute to the development of a compact physics-based system in the form of (see \citep{Vogiatzoglou2024} for details):
\begin{align}
\label{eq:EnergyBalanceDimlessModel}
\frac{\partial T}{\partial t} &= \alpha_1 \Big(\underbrace{\mathbf{D} \boldsymbol{\nabla}^2_{xy} T}_{\text{dispersion}} -\underbrace{\mathbf{u} \cdot \boldsymbol{\nabla}_{xy} T}_{\text{advection}} \Big) -
\underbrace{\alpha_2 E r_{\rm e} + \alpha_3 X r_{\rm x}}_{\text{reaction}}  - \ \alpha_4 \underbrace{\textnormal{U} \left(T - T_{\rm a}\right)}_{\text{convection}}, \\
\label{eq:Reaction1Model}
\frac{\partial E}{\partial t} &=- E r_{\rm e},\\
\label{eq:Reaction2Model}
\frac{\partial X}{\partial t} &=-  X r_{\rm x},
\end{align}
where $T$ denotes once more the temperature of the fire layer, $E$ the endothermic fuel mass fraction, and $X$ the exothermic fuel mass fraction. The coefficients $\alpha_1, \alpha_2, \alpha_3$, and $\alpha_4$ integrate fuel, combustion, and environmental properties, while the notation $T_{\rm a}$ ([=] K) denotes the ambient temperature. Moreover,  $\mathbf{D} = [\textnormal{D}_x\,\, \textnormal{D}_y]^\T$ ([=] m$^2$/s) is the dispersion coefficient vector, $\mathbf{u} = [\textnormal{u}_x\,\, \textnormal{u}_y]^\T$ ([=] m/s) is the velocity vector replicating gas flow averaged over the plantation height, and $\textnormal{U}$ ([=] W/m$^2$K) is the overall heat transfer coefficient governing thermal losses to the environment.

\subsection{Parameter learning}
\label{sec:Parameter_Learning}

The spread of forest fires is driven by highly non-linear natural events, influenced by numerous dynamically competitive subprocesses and multi-physics phenomena that appear to affect the intensity and direction of combustion fronts. The envisioned physics-based model employs a straightforward mathematical concept to incorporate these attributes and sufficiently simulate wildfire expansion, relying on a limited set of crucial model parameters articulated by explicit physical terms. Therefore, within the context of parameter inference, this work aims to elucidate the substantial impact of $\mathbf{D}, \mathbf{u}$, and $\textnormal{U}$ parameters, compactly represented as $\bm{\theta} = [\mathbf{D}^\T\,\, \mathbf{u}^\T\,\,\textnormal{U}]^\T$, on the rate of firefront spread \citep{Prieto2017, Nelson2015}.

It is generally accepted in the literature that, under different prevailing weather conditions, radiation or convection have a dominant effect on the shape of the flame front and its spreading characteristics \citep{FrankmanD2013}. While convection tends to dominate wind-driven propagation (eruptive fires) by advancing the hot combustible gases directly to the unburned fuel, radiation primarily enhances the immediate preheating and ignition of fuel (plume-dominated fires) \citep{Simeoni2011}. The dispersion coefficient, $\mathbf{D}$, of the present model integrates the combined effects of radiation, buoyant instabilities, and turbulent streamwise vortices. As an uncertain model parameter, it also depends on the dynamic fluctuation of the mean gaseous velocity \citep{Finney2015} and the characteristic length scales of the firefront. Specifically, this coefficient combines the fireline width \citep{Cussler}, measured along the flame axis, and the fireline length \citep{CheneyNP1995}, measured from the head to the lateral flame parts. A shorter fireline favors more efficient re-direction of wind around the main fire spot, while a longer fireline tends to force wind penetration through the firefront. As a result, the firefront spreads faster forward and also extends progressively in the lateral direction (creating the well-known parabolic firefront shape) \citep{JMCanfield2014, Nelson2015}. Thus, this parameter is of profound importance in shaping the fireline geometry and influencing the acceleration or deceleration of the growth and spread rate.

The influence of wind is a critical factor in wildfire behavior \citep{Finnigan2007, Inoue1963}, dictating not only the ROS but also the overall dynamics of fire propagation. However, wind measurements remain susceptible to significant sparsity and variability stemming from abrupt spatial and temporal atmospheric and topographical irregularities \citep{Forthofer2014}. The mean gaseous velocity, $\mathbf{u}$, fluctuates between minimum values due to the drag resistance of the foliage and upward plume (vertical wind shear stresses), reaching maximum values when the canopy is diminished by fire spread. Furthermore, gas flow impacts both the dispersion and convection heat terms, which in turn influence the magnitude of heat fluxes transferred to the ambient \citep{JMCanfield2014}.

The overall heat transfer coefficient, $\textnormal{U}$, holds significant importance as it signifies the extent of heat release to the surroundings. This stochastic term in the physical model accounts for the combined effects of both natural convection \citep{Murgai1960} and radiation \citep{Simeoni2011, Finney2013} and typically constitutes a non-linear function of temperature, as defined by both the Boltzmann law and the Nusselt correlation (for free convection). When this coefficient increases, heat losses intensify, leading to a passive combustion front and, thus, a halt in fire propagation. On the contrary, lower values enhance the fire field with higher temperature records and intense progression rates.

In the upcoming case studies, the examination will focus on a flat terrain where the gaseous phase moves uniformly through and above the plantation. This simplification of both topography, fuel models, and wind flow, even in the real-world case study of the Troy Fire, is essential to demonstrate the ability of PiNNs to approximate fundamental physical coefficients without explicit knowledge of the topology and fuel distribution, a common limitation in actual scenarios \citep{Heui2012}. The primary objective remains the effective application of ANNs for parameter learning from a quantitative perspective under emergent circumstances in the presence of considerable uncertainty, rather than an exhaustive exploration of the complex physics of terrain, wind, and fuel modeling. However, the incorporation of terrain inclination and sophisticated fuel models,  such as those based on Rothermel’s surface fire spread correlation \citep{ScottBurgan2005}, is currently under development in our modeling formulation.

The aforementioned model parameters appear to be the most appropriate for learning purposes, as they account for a substantial portion of both model and environmental uncertainties, ultimately exerting a profound impact on the dynamics of forest fires. Further details concerning model derivation and parameter descriptions can be found in \citep{Vogiatzoglou2024}. To effectively train physics-informed neural networks for parameter inference, the synthetic data used as training sets in \Cref{sec:Application_and_Results} employs either constant values for these coefficients or values that fluctuate over time to account for the effect of model error in the inference process. In the latter case, the fluctuations are considered to be samples of a Gaussian process with a specific correlation structure. Finally, the Troy Fire case study is examined to evaluate the PiNN’s ability to learn the considered wildfire model parameters under a real-world fire scenario.

\section{Physics-informed neural networks for inverse optimization}
\label{sec:PiNNs}

\subsection{Artificial neural networks}
\label{sec:Architecture}

This section briefly describes the architecture of artificial neural networks and their primitive functionality. Consider a fully-connected feed-forward neural network represented by a compositional function, ${\cal N}^{L}(\mathbf{x}): \R^{n} \to \R^{m}$, of $L$ layers, while $N_{\ell}$ indicates the number of artificial neurons at each intermediate (hidden) layer, with $\ell =$ 1, 2, \ldots, $L - 1$. In terms of the initial (input) layer, $N_{0} = n$, reflecting the dimension of the input training data, $\mathbf{x} \in \R^{n}$. In contrast, the final (output) layer has $N_{L} = m$, determined by the dimension of the output quantities of interest. Commonly, ANNs are tailored for either classification tasks, such as organizing data into specific categories, or regression tasks, like predicting numerical values given a set of training data points.

Every neuron's output passes through an activation function (AF), $\sigma(\mathbf{x}): \R^{n} \to \R^{n}$, which is an operator introducing non-linearity to the network, enabling it to capture intricate patterns present in the training dataset. The activation function transforms the weighted linear sum input of one layer into a non-linear output for the subsequent layer and determines whether a neuron is excited or inhibited. Several activation functions are available, among which the most common are the logistic sigmoid, $\sigma(x) = \frac{1}{1 + e^{-x}}$, the hyperbolic tangent, $\sigma(x) = \tanh(x) =  \frac{e^{x} - e^{-x}}{e^{x} + e^{-x}}$, and the rectified linear unit (ReLU), $\sigma(x) =$ max($x$, 0), which can be approximated by a smooth differentiable function of the form $\sigma(x)  = \ln(1 + e^x)$. In general, activation functions have the property: $-\infty < \lim_{x \to -\infty} \sigma(x) <  \lim_{x\to \infty} \sigma(x) < \infty$.

\begin{figure}[tbp]
\centering
\includegraphics[width=\textwidth]{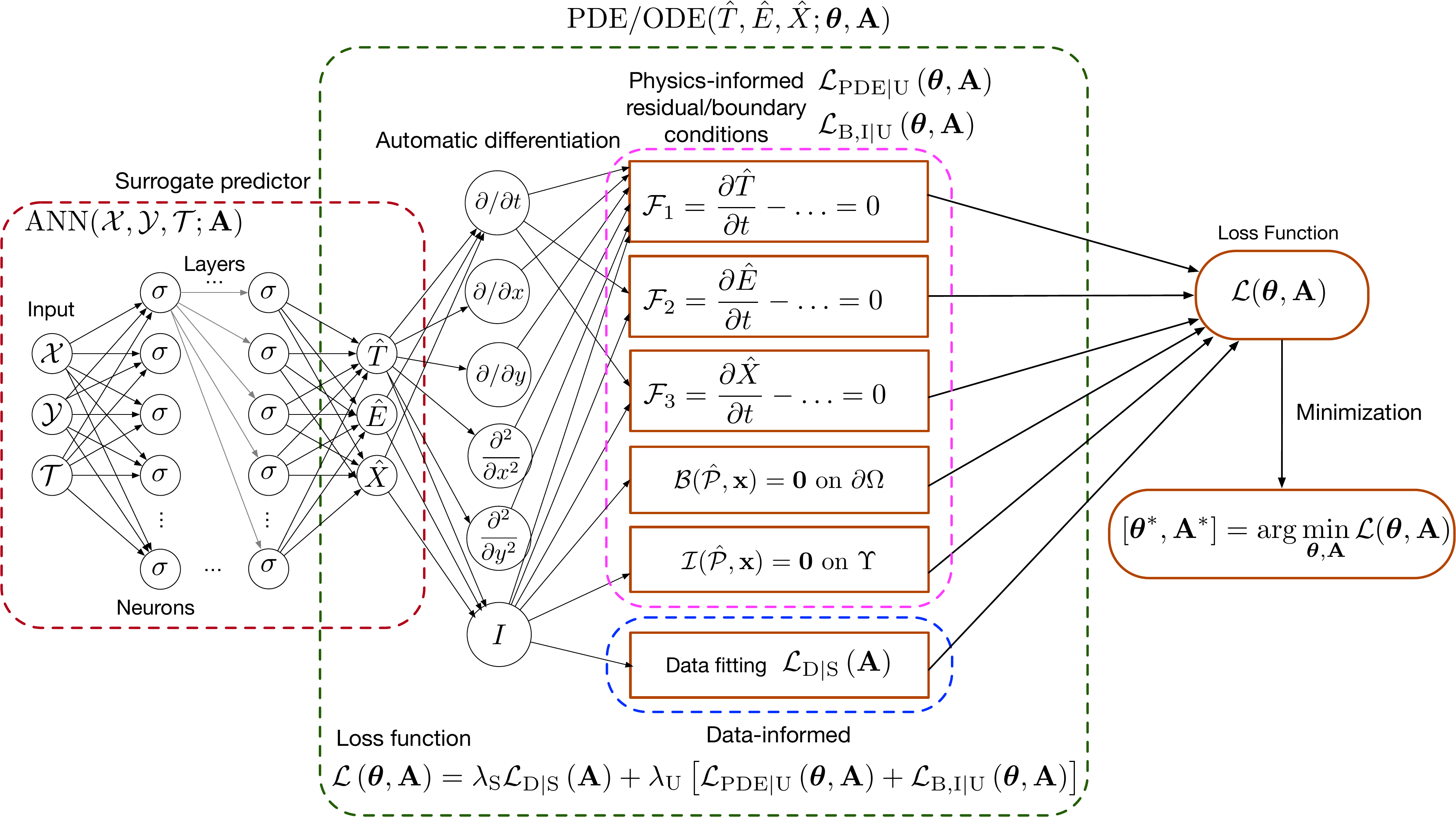}
\caption{Architecture of the proposed PiNN for learning the parameters of the wildfire spreading model. The ANN on the left represents the physics-uninformed (data-driven) surrogate predictor, while the right network illustrates the physics-informed (data-uninformed though) residual, initial and boundary conditions of the wildfire model, and the data-informed cost function (data fitting) that penalizes deviation of the surrogate model predictions from the synthetic data.} 
\label{fig:PiNN_Diagram}
\end{figure}

A neural network can be understood as a function approximator or surrogate model parameterized by weights, which adjust the strength of connections between neurons of different layers, and biases, which control the flexibility of the network. Let $\W^{\ell} \in \R^{N_{\ell}\times N_{\ell-1}}$ and $\b^{\ell} \in \R^{N_{\ell}}$ be the weight matrix and bias vector at the $\ell$-th layer, where $\ell = 1, 2,..., L$, respectively. The primary objective of this approach involves fixing the weights and biases, or the concatenated vector, $\mathbf{A} = \{\W^{\ell}, \, \b^{\ell}\}_{1 \leq\ell \leq L}$, to their optimal values, leveraging labeled pairs of input and output data (supervised learning). Specifically, the ANN tries to minimize the discrepancy between predicted and target outputs via a loss function and enhance its generalization ability to unseen data. Regarding this, automatic (or algorithmic) differentiation \citep{Baydin2018} in general and the backpropagation algorithm \citep{Rumelhart1986} in particular contributes to the efficient computation of loss gradients with respect to the network parameters. Moreover, for the nonlinear, high-dimensional optimization problem present, gradient-based optimizers and their variants (e.g., stochastic gradient descent (SGD) \citep{Robbins1951}) as well as Adam \citep{Kingma2014} and L-BFGS \citep{Byrd1995} algorithms are commonly employed to search for global minima. The architecture stated above can be described as follows (see also the left part of \Cref{fig:PiNN_Diagram}):
\begin{align}
\label{eq:NN_Arch1}
\text{Input layer:} &\quad {\cal N}^0(\mathbf{x}) = \mathbf{x} \in \R^n,\\ 
\label{eq:NN_Arch2}
\text{Hidden layers:} &\quad  {\cal N}^{\ell}(\mathbf{x}) = \sigma(\W^{\ell}{\cal N}^{\ell - 1}(\mathbf{x}) + \b^{\ell}) \in \R^{N_{\ell}} \quad \text{for} \quad \ell = 1, 2, \ldots, L - 1, \\ 
\label{eq:NN_Arch3}
\text{Output layer:} &\quad  {\cal N}^{L}(\mathbf{x}) = \W^{L}{\cal N}^{L - 1}(\mathbf{x}) + \b^{L} \in \R^{m}.
\end{align}

The Universal Approximation Theorem (UAT), the most important theoretical result in neural networks (NNs), states that a neural network of the architecture \eqref{eq:NN_Arch1}--\eqref{eq:NN_Arch3} and a large class of activation functions (including sigmoid and ReLU) can approximate arbitrarily closely (in an appropriate mathematical sense) any piecewise continuous function provided the number of hidden layers is sufficiently large, showing that universal approximation is essentially implied by the network structure  \citep{Cybenko1989, Funahashi1989, Hornik1989, Leshno1993}. While the UAT provides some assurance about the capability of ANNs to approximate functions to an arbitrary degree of accuracy, it does not provide any rigorous theoretical results on the size of the ANN (in terms of hidden layers and neurons per hidden layer) needed nor its performance. In practice, a trial-and-error procedure can be pursued to obtain an appropriate set of hyperparameters that result in "good" performance (accuracy of approximation). For a more systematic hyperparameter search to tune deep ANNs, Bayesian optimization can be employed, as in AlphaZero \citep{Silver2017}.

The contribution of PiNNs to high-dimensional contexts governed by parameterized equations has recently surfaced, stepping into both forward (data-driven prediction) and inverse (data-driven discovery) problems with great promises. The integration of PDEs and/or ODEs for regularization purposes by enforcing the physical laws of the dynamic system (prior knowledge) in accordance with the selected training dataset signifies a substantial advancement towards improving the performance of the learning algorithm (accelerated convergence).  

\subsection{Training of PINNs}
\label{sec:Training}

Consider the wildfire spreading model~\eqref{eq:EnergyBalanceDimlessModel}--\eqref{eq:Reaction2Model}, with an explicit solution denoted by ${\cal P} \equiv {\cal P}(\mathbf{x}) = [T(\mathbf{x}) \  E(\mathbf{x}) \ X(\mathbf{x})]^\T$, defined over the spatial domain, $\Omega \subset \R^2$, and temporal domain, $\Upsilon$, where $\mathbf{x} \in \Omega \times \Upsilon$ is the input, parameterized by the vector, $\bm{\theta} = [\mathbf{D}^\T\,\, \mathbf{u}^\T\,\,\textnormal{U}]^\T$ (to be learned). The residuals of \eqref{eq:EnergyBalanceDimlessModel}--\eqref{eq:Reaction2Model} are given by:
\begin{align}
\label{eq:Residual_1}
{\cal F}_{1}(T_t, T_x, T_y, T_{xx}, T_{yy}, {\cal P}, \bm{\theta}) &= \frac{\partial T}{\partial t} - \alpha_1 \Big(\mathbf{D} \boldsymbol{\nabla}^2_{xy} T - \mathbf{u} \cdot \boldsymbol{\nabla}_{xy} T \Big) +
\alpha_2 E r_{\rm e} - \alpha_3 X r_{\rm x} + \alpha_4 \textnormal{U} (T - T_{\rm a}) = 0,\\
\label{eq:Residual_2}
{\cal F}_{2}(E_{t}, {\cal P}) &= \frac{\partial E}{\partial t} + E r_{\rm e} = 0, \\
\label{eq:Resisual_3}
{\cal F}_{3}(X_{t}, {\cal P}) &= \frac{\partial X}{\partial t} + X r_{\rm x} = 0. 
\end{align}
The mathematical system described above is imposed by a set of boundary conditions, ${\cal B}({\cal P}, \mathbf{x})$, on $\partial\Omega$ and initial conditions, ${\cal I}({\cal P},  \mathbf{x})$, on $\Upsilon$, emulating the physics driving firefronts. Taking advantage of a training dataset, ${\cal D} = \{{\cal P}(\mathbf{x})\}_{\mathbf{x} \in \Omega \times \Upsilon}$, including temperature and fuel measurements, the neural network is constrained to generate predictions satisfying both the physical model (eqs.~\eqref{eq:Residual_1}--\eqref{eq:Resisual_3}) and the boundary conditions, ${\cal B}({\cal P}, \mathbf{x}) = \bf 0$, as well as the initial conditions, ${\cal I}({\cal P}, \mathbf{x}) = \bf 0$. This is particularly essential during a real-world fire event, as it yields accurate estimates grounded in the underlying physics (see \Cref{sec:Case_5}).

The input layer incorporates labeled data with specific spatial and temporal sampling frequencies. Spatial points are denoted as ${\cal X} = \{x_1, \  x_2, \  ..., \  x_{|{\cal X}|}\}$, ${\cal Y} = \{y_1, \  y_2, \  ..., \  y_{|{\cal Y}|}\}$ $\in \Omega$, and temporal points as ${\cal T} = \{t_1, \  t_2, \  ..., \  t_{|{\cal T}|}\}$ $\in \Upsilon$, where $|{\cal X}|, |{\cal Y}|, |{\cal T}|$ express the cardinality of sets ${\cal X}$, ${\cal Y}$, and ${\cal T}$, respectively. The selection of the appropriate quantity of input data is paramount, necessitating the inclusion of sufficient points from the boundaries ($\in \partial\Omega$) and initial states ($\in \Upsilon$). The input layer equipped with the aforementioned datasets is generic, thus other wildfire data (e.g., thermal images from drones or planes equipped with thermal cameras) and meteorological data (e.g., climate and weather, wind speed, etc.) can be readily incorporated as input data, see e.g., \citep{Dabrowski2023, Shadrin2024}. During training, the neural network is fed by such input pairs of ${\cal X}, {\cal Y}, {\cal T}$ from the entire training spatiotemporal set, and by integrating data points, ${\cal D} = \{{\cal P}(\mathbf{x})\}_{\mathbf{x} \in \Omega \times \Upsilon}$, attempts to predict the state variables, $\hat{{\cal P}} = [\hat{T} \  \hat{E} \ \hat{X}]^\T$. To evaluate the efficiency of the network to match predictions with the target values, a mean squared error (MSE) loss function ($L_2$-norm) is employed. This total loss function, ${\cal L}$, includes three critical components: (i) a supervised (data-driven) cost criterion, ${\cal L}_{\rm D|S}$, determined by measurements ${\cal P}$ and predictions $\hat{{\cal P}}$, at multiple spatiotemporal points; (ii) an unsupervised loss function, ${\cal L}_{\rm PDE|U}$, for the set of residual PDEs; and (iii) an unsupervised loss function, ${\cal L}_{\rm B,I|U}$, for the boundary (${\cal B}$) and initial (${\cal I}$) conditions. Thus, the total loss function is expressed as:
\begin{equation}
\label{eq:Loss_function}
{\cal L}\left(\bm{\theta}, \mathbf{A}\right)= \lambda_{\rm S}{\cal L}_{\rm D|S}\left(\mathbf{A}\right) +  \lambda_{\rm U}\left[{\cal L}_{\rm PDE|U}\left(\bm{\theta}, \mathbf{A}\right) + {\cal L}_{\rm B,I|U}\left(\bm{\theta}, \mathbf{A}\right)\right],
\end{equation}
with,
\begin{align}
\label{eq:Loss_1}
{\cal L}_{\rm D|S}\left(\mathbf{A}\right) &= \frac{1}{N_D}\sum_{i=1}^{N_D}\|{\cal P} - \hat{{\cal P}}\|^2, \\
\label{eq:Loss_2}
{\cal L}_{\rm PDE|U}\left(\bm{\theta}, \mathbf{A}\right) &= \frac{1}{N_{PDE}}\sum_{i=1}^{N_{PDE}}\left(\|{\cal F}_{1}(\hat{T_t}, \hat{T}_x, \hat{T_y}, \hat{T}_{xx}, \hat{T}_{yy}, \hat{{\cal P}}, \bm{\theta})\|^2 + \|{\cal F}_{2}(\hat{E_{t}}, \hat{{\cal P}})\|^2 + \|{\cal F}_{3}(\hat{X_{t}}, \hat{{\cal P}})\|^2\right), \\
\label{eq:Loss_3}
{\cal L}_{\rm B,I|U}\left(\bm{\theta}, \mathbf{A}\right) &= \frac{1}{N_{B,I}}\sum_{i=1}^{N_{B,I}}\left(\|{\cal B}(\hat{{\cal P}}, \mathbf{x})\|^2 + \|{\cal I}(\hat{{\cal P}}, \mathbf{x})\|^2\right), 
\end{align}
where $\lambda_{\rm S}$ and $\lambda_{\rm U}$ are non-negative penalty terms (loss weights) for the supervised and unsupervised loss functions, respectively, which balance the contribution of each sub-loss term. In our case studies, these values are set to 1.0. Although extensive experiments have been conducted (across all case studies presented herein) using various values for these weights, the final results demonstrate the robustness of PiNNs in parameter identification without overemphasizing any specific component of the problem. These weights help guide the optimization process by smoothing the loss landscape, leading to improved convergence and stability during training. Overall, our selection ensures that the model effectively learns the underlying physical laws while appropriately fitting the observed data and boundary-initial conditions. Additionally, the selection of these two penalty terms can be either carried out via a trial-and-error procedure (user-defined) or tuned automatically to speed up convergence. The notations $N_D, N_{PDE}$, and $N_{B,I}$ correspond to the total number of sampling points for each part of the loss function.

In summary, PiNNs typically consist of: (i) a conventional multi-layer neural network (physics-uninformed predictor); (ii) the residual constraints of the dynamic system (physics-informed part), simultaneously accounting for both boundary and initial conditions; and (iii) a total loss function (data-driven fitting cost criterion). This seamless architecture connecting these components is illustrated in the right portion of \Cref{fig:PiNN_Diagram} and is applied identically in both the simulated and real case studies. The ultimate objective (under the concept of the inverse optimization problem) is to learn the best-fit values for both the unknown model parameters, denoted as $\bm{\theta}^{*} = [\mathbf{D}^{{*}^\T}\,\, \mathbf{u}^{{*}\T}\,\,\textnormal{U}^{{*}}]^\T$, and the weights and biases, denoted as $\mathbf{A}^{*} = \{\W^{\ell^*}, \, \b^{\ell^*}\}_{1 \leq\ell \leq L}$, by minimizing the total loss function  \eqref{eq:Loss_function} during training. This can be formulated as:
\begin{equation}
\label{eq:Optimization}
[\bm{\theta}^{*}, \mathbf{A}^{*}] = \arg \min_{\bm{\theta}, \mathbf{A}} {\cal L}\left(\bm{\theta}, \mathbf{A}\right).
\end{equation}
The underlying optimization problem is highly non-linear and non-convex within a high-dimensional parameter space, provided that the activation functions in each neuron of the ANN are nonlinear operators. However, it can be readily solved through algorithmic differentiation and the employment of batch methods, such as SGD, Adam, or L-BFGS, the latter leverages Hessian information.

\section{Application and results}
\label{sec:Application_and_Results}

\subsection{Parameter learning setup}
\label{sec:Parameter_Learning_Setup}

This section underscores the effectiveness of PiNNs in fine-tuning the unknown parameters of the wildfire spreading model \eqref{eq:EnergyBalanceDimlessModel}--\eqref{eq:Reaction2Model}. Specifically, five representative case studies are conducted: \emph{Case Study 1} explores the ability of PiNNs to learn the parameters for the spatiotemporal evolution of a one-dimensional (1D) firefront of the wildfire spreading model; \emph{Case Study 2} seeks to train neural network to adjust the same model parameters to their optimal values while dealing with synthetic noisy data; \emph{Case Study 3} and \emph{Case Study 4} demonstrate the predictive ability of PiNNs in learning the parameters for the spatiotemporal evolution of a two-dimensional (2D) firefront on a plane surface without and with noisy data, respectively. Finally, \emph{Case Study 5} investigates the firefront evolution of the Troy Fire in California, 2002, aiming to identify the unknown model parameters based on an actual wildfire event.

It is important to emphasize that most of the aforementioned case studies are carried out on flat terrain with a uniform fuel composition (except for \emph{Case Study 5}) and consistent wind direction across the domain. Moreover, both the dispersion coefficient, $\mathbf{D}$, and the velocity term, $\mathbf{u}$, indicate vector quantities with components in both the streamwise ($x-$) and spanwise ($y-$) directions, respectively. Finally, the wildfire spreading model \eqref{eq:EnergyBalanceDimlessModel}--\eqref{eq:Reaction2Model} has been converted into a dimensionless form throughout the entirety of this paper (as for the efficient functionality of the network), allowing for the utilization of non-dimensional simulated temperature and fuel measurements with respect to the training dataset.  
\subsection{PiNN for 1D firefront}
\label{sec:1D_modeling}

For both \emph{Case Study 1} (see \Cref{sec:Case_1}) and \emph{Case Study 2} (see \Cref{sec:Case_2}), the neural network architecture consists of one (1) input layer containing two (2) neurons, representing the discretized spatiotemporal domain $\in \R^2$ ($x \in \Omega$ for spatial and $t \in \Upsilon$ for temporal coordinates). This is followed by four (4) hidden layers, each with twenty (20) neurons, and finally concludes with one (1) output layer with three (3) neurons, which serve to provide the predicted state variables, $\hat{{\cal P}} = [\hat{T} \  \hat{E} \ \hat{X}]^\T \in \R^3$. Given the notation for the NN's architecture (see \Cref{sec:Architecture}): $L$ = 5, $N_{0}$ = 2, $N_{\ell}$ = 20 for $\ell =$ 1, 2, 3, 4, and
$N_{5}$ = 3. Thus, the parameterized set of weights and biases can be expressed as $\mathbf{A} = \left\{\W^{1}, \W^{2}, \W^{3},\W^{4}, \W^{5}, \b^{1}, \b^{2}, \b^{3}, \b^{4}, \b^{5}\right\} \in \R^{1383}$. This neural network architecture was developed based on an independent analysis of the optimal number of neurons and hidden layers. To confirm its "ideal" structure, various configurations with different combinations of neurons and layers were tested and evaluated based on performance metrics like accuracy and computational efficiency. The selected architecture showed the best trade-off between these metrics, making it the most effective choice for the identification task.

The output of each neuron (in terms of hidden and output layers) undergoes a logistic sigmoid activation function, $\sigma(x) = \frac{1}{1 + e^{-x}}$, enabling the network's capacity to discern intricate non-linear correlations within the training dataset. The logistic sigmoid activation function offers a smooth gradient, ensuring stable weight updates during the training process. Moreover, its output is strictly non-negative (output values in the range from 0 to 1), which is particularly important for the temperature component, as it prevents the model from producing physically implausible negative values, unlike the hyperbolic tangent AF. To expedite convergence, a hybrid training scheme is adopted. Initially, the Adam algorithm is employed for 40.000 iterations, after which the optimization process transitions to L-BFGS for the remainder of the training. The learning rate is set to 0.0003 to induce smooth updates during the exploration stage. This adjustment is essential for achieving improved performance and accelerated convergence. The training is conducted on a computer with an AMD Ryzen Threadripper PRO 3995WX at 2.70 GHz and 32 cores, operating on Windows 10 Pro 64-bit.

The wildfire spreading model \eqref{eq:EnergyBalanceDimlessModel}--\eqref{eq:Reaction2Model} is governed by a simple set of boundary and initial conditions. Specifically, at the initial state ($t$ = 0), the conditions, ${\cal I}$, are set as follows: $T(x, 0) = T_{p}e^{-\frac{(x - x_0)^2}{\gamma^2}} + T_{\rm a}, \,  E(x, 0) = E_{0}, \, $ and $X(x, 0) = X_{0}$ $\forall \, x \in \Omega$. Here, $T_{p}$ ([=] K) represents an approximation of the pyrolysis temperature, $x_0$ ([=] m) denotes the spatial coordinate corresponding to the maximum temperature value during the ignition phase, $\gamma$ ([=] m) signifies the extend of the initial fire spot, and $E_{0}, X_{0}$ denote the constant compositions of the endothermic and exothermic fuel, respectively. Furthermore, to account for boundaries unaffected by fire progression (satisfactorily extended domain), Dirichlet boundary conditions, ${\cal B}$, are employed, wherein: $T(0, t) = T(L, t) = T_{\rm a}, \, E(0, t) = E(L, t) = E_{0},$ and $X(0, t) = X(L, t) = X_{0}$ $\forall \, t \in \Upsilon$ with $L$ ([=] m) characterizing the total length of the simulated domain.

\subsubsection{Parameter learning for 1D firefront with synthetic data}
\label{sec:Case_1}

The main focus of \emph{Case Study 1} is to demonstrate the pivotal role of PiNNs in inferring the three unknown physical quantities, $\textnormal{D}, \textnormal{u}$, and $\textnormal{U}$ (scalar values), within the wildfire spreading model. Synthetic data are meticulously prepared and tested applying known values for the above model parameters, denoted as $\bm{\theta}^{*} = [\textnormal{D}^{*}\,\, \textnormal{u}^{*}\,\,\textnormal{U}^{*}]^\T = \left[0.41\,\, 0.25\,\, 0.61\right]^\T$. The data generation process (i.e., generated via a physics-informed digital twin) involves explicitly solving the mathematical system of equations~\eqref{eq:EnergyBalanceDimlessModel}--\eqref{eq:Reaction2Model} through finite difference numerical methods (i.e., central difference scheme for the discretization of the diffusion term and upwinding scheme for the discretization of the advection term) while ensuring compliance to the prescribed boundary and initial conditions. As the application and significance of the proposed approach are inherently real-world, we implemented the Adaptive Mesh Refinement (AMR) numerical scheme to reduce computational demands to manageable levels, achieving 0.1 CPU-seconds of computation time per 1 CPU-second of simulated time \citep{Chatzimanolakis2022a, Vogiatzoglou2024}. The simulation covers a horizontal spatial domain of $x \in \left[0, 100\right]$ m and a time interval of $t \in \left[0, 200\right]$ s, while the training dataset includes points sampled at consecutive uniform intervals of $d_x =$ 1 m in space and $d_t =$ 1 s in time (equivalent to a frequency of 1 Hz) with a total number of 20.301 data points for each state variable ($T, E, X$).

\begin{figure}[tbp]
\centering
\includegraphics[width=.55\textwidth]{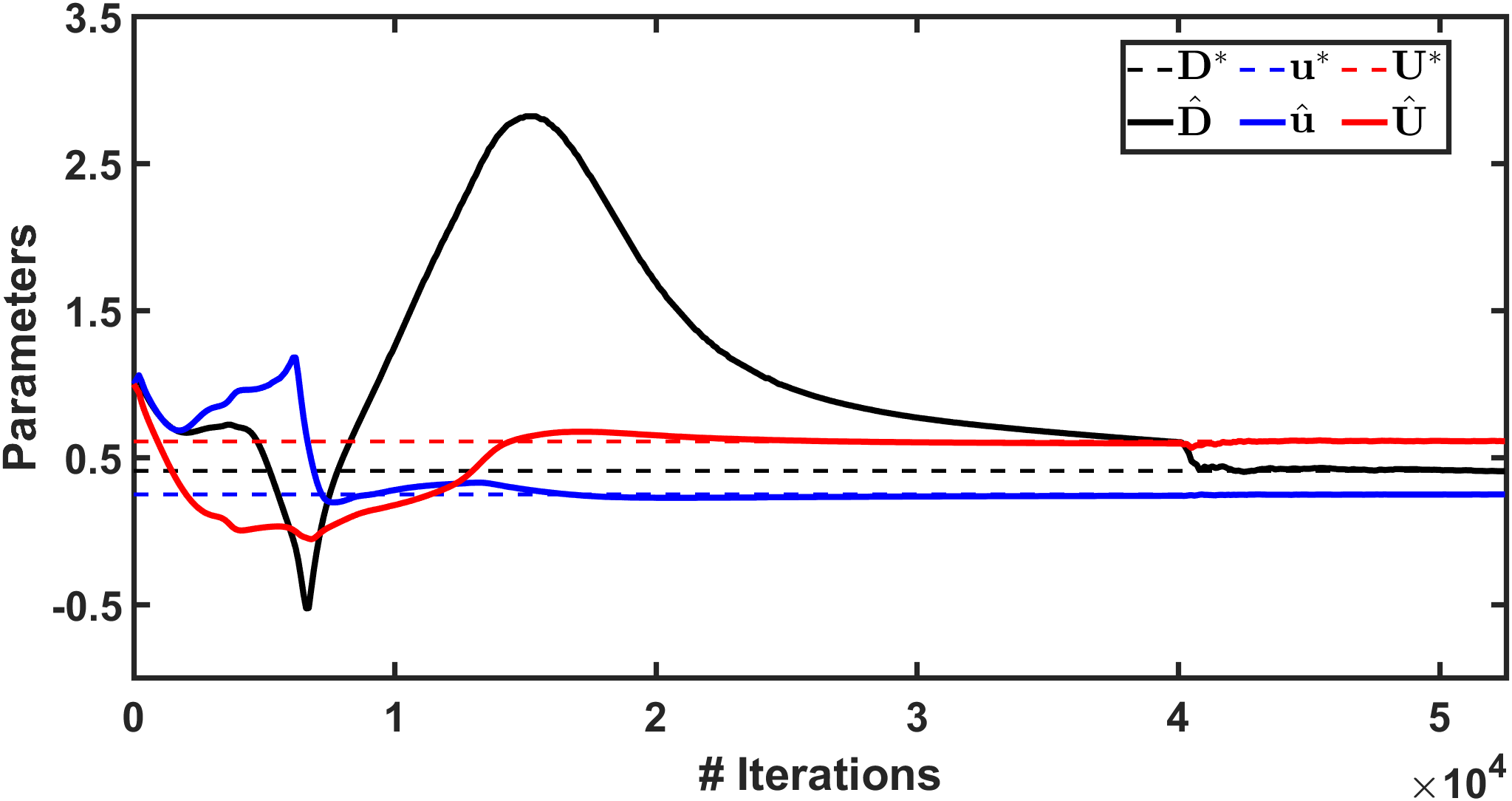}
\caption{Parameter learning and convergence in the 1D firefront of the wildfire spreading model. The predicted vector of the three model parameters is $\hat{\bm{\theta}} = \left[0.408\,\, 0.25\,\, 0.61\right]^\T$, while the true vector used for generating the training dataset is $\bm{\theta}^{*} = \left[0.41\,\, 0.25\,\, 0.61\right]^\T$.}
\label{fig:Parameter_learning_1}
\end{figure}

The outcomes exemplify the efficacy of a well-employed data-assimilation methodology. The ANN implemented the synthetic data during the training part to approximate the a priori known model parameters, $\bm{\theta}^{*}$, while initializing all the unknown coefficients with the estimate 1.0 (during the first iteration). The predicted values have ultimately converged to their nominal counterparts, yielding $\hat{\bm{\theta}} = [\hat{\textnormal{D}}\,\, \hat{\textnormal{u}}\,\,\hat{\textnormal{U}}]^\T = \left[0.408\,\, 0.25\,\, 0.61\right]^\T  \approx \bm{\theta}^{*}$. \Cref{fig:Parameter_learning_1} illustrates the convergence behavior of the unknown model parameters over the total course of training iterations. Notably, coefficients $\textnormal{u}$ and  $\textnormal{U}$ exhibit significant convergence after 20.000 iterations, while the L-BFGS optimizer was instrumental in achieving complete alignment beyond 40.000 iterations. The effectiveness of PiNNs is further demonstrated by their robustness in managing circumstances where dispersion and convection coefficients stepped into negative ranges during the exploration stage. The network exploited the embedded prior knowledge (physical modeling constraints) and adeptly adjusted its "learning direction" towards positive ones.

\begin{figure}[tbp]
\centering 
\includegraphics[width=\textwidth]{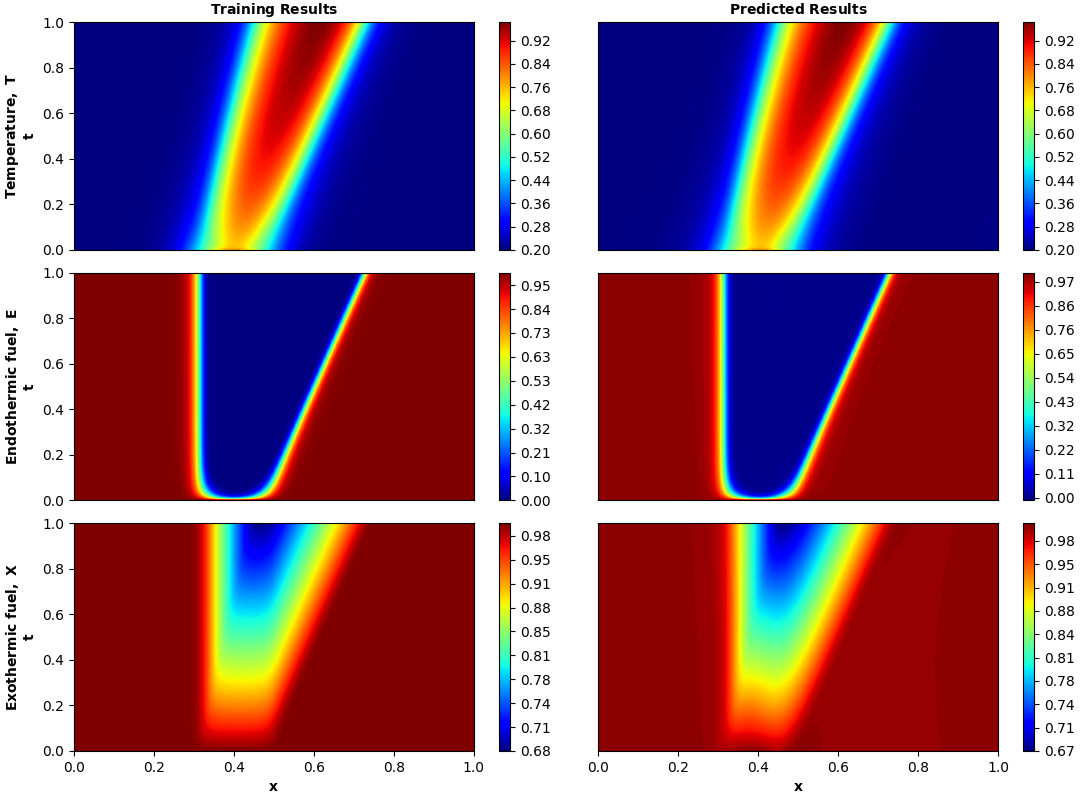}
\caption{1D spatiotemporal firefront for the three state variables: $T, E, X$. The left column presents the explicit solution of the physics-based wildfire spreading model, while the right column displays the prediction of PINNs. Both cases are provided in a dimensionless form.}
\label{fig:Comparison_Matlab_PiNNs}
\end{figure}

\begin{figure}[tbp]
\centering
\begin{tabular}{c c} \includegraphics[width=0.42\textwidth]{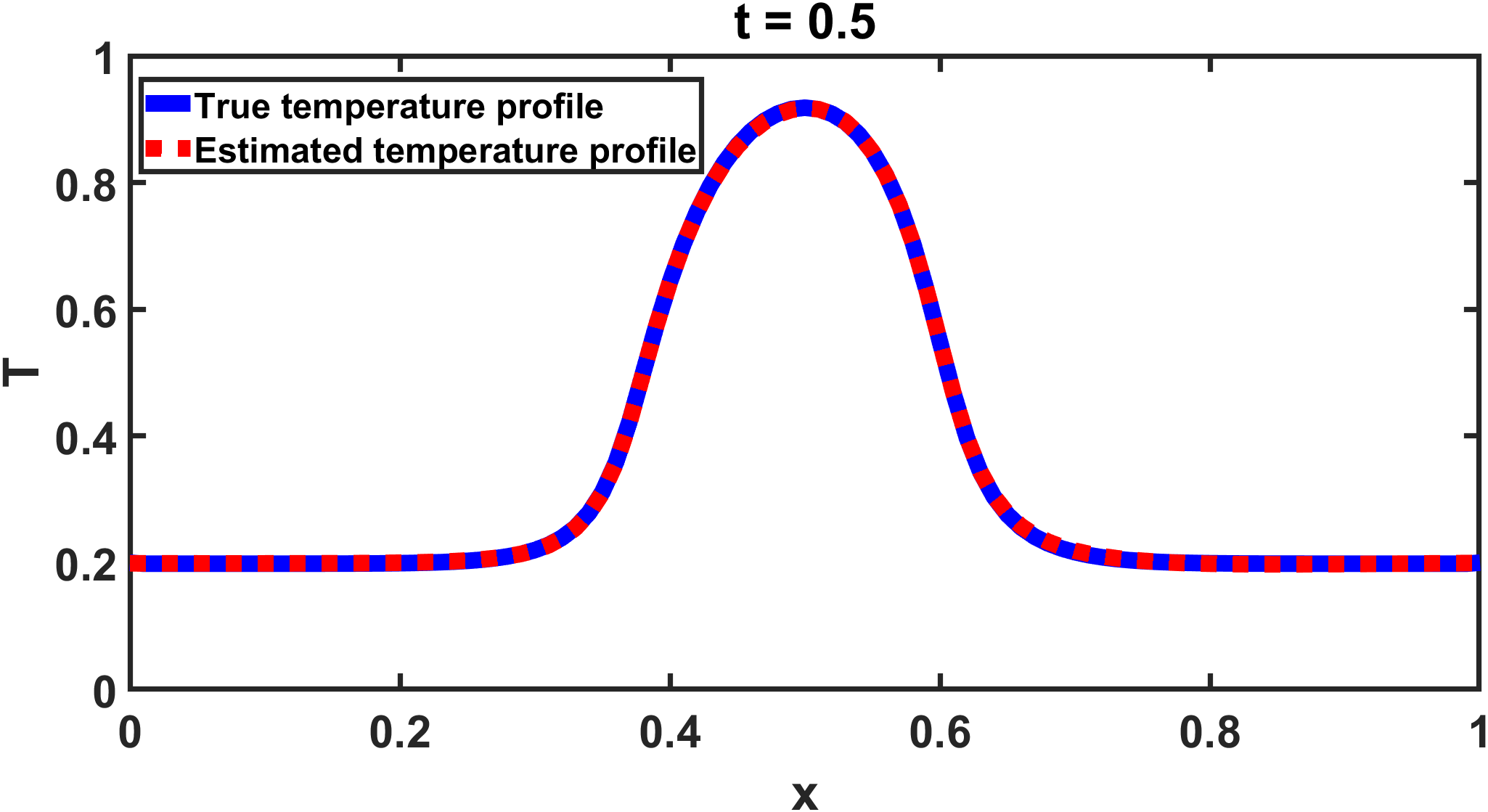} &
\includegraphics[width=0.42\textwidth]{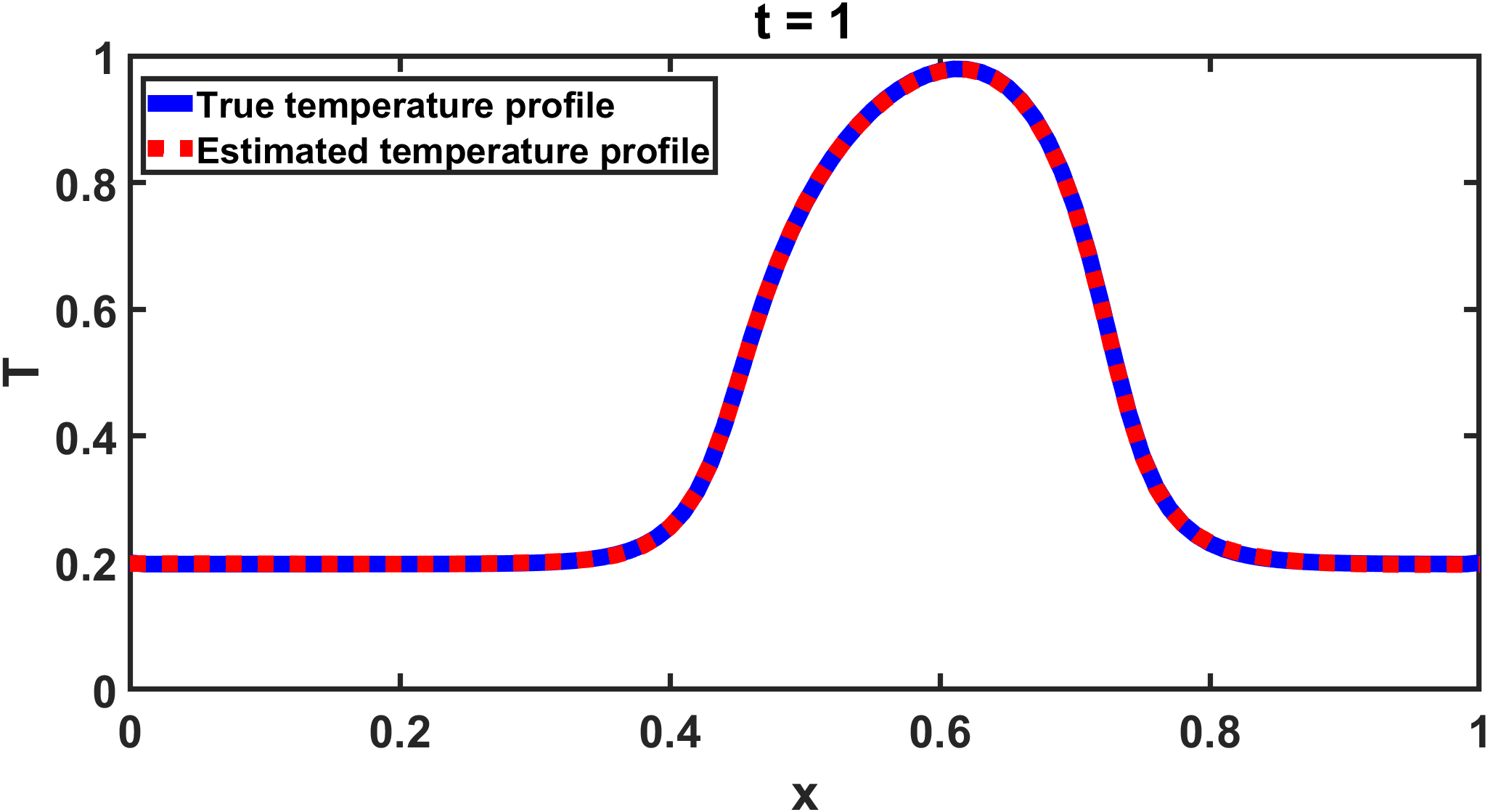}
\end{tabular}
\caption{Comparison of the temperature distribution between the explicit solution and the prediction of PiNNs at two specific non-dimensional time instants: $t = 0.5$ and $t =1$, for all nodes across the spatial firefront.}
\label{fig:1D_time_comparison}
\end{figure}

With these final estimations, the neural network is calibrated to forecast the entire fire evolution domain across both space ($x$) and time ($t$). \Cref{fig:Comparison_Matlab_PiNNs} provides a comparison between the explicit solution (analytically solved) of the physics-based mathematical system and the prediction delivered by PiNNs, demonstrating their ability to accurately replicate the target dataset with minimal expenses in the hidden layer architecture \citep{Raissi2018}. To enhance clarity, a detailed comparison between the nominal and predicted temperature distributions is conducted at two specific non-dimensional time instants, $t =$ 0.5 (intermediate) and 1 (final), for all nodes across the domain. This comparison is delineated in \Cref{fig:1D_time_comparison}, exhibiting the precise alignment between the true and estimated temperature profiles. All these findings demonstrate the substantial predictive capabilities of PiNNs, affirming their effectiveness in accurately forecasting temperature and fuel adjustments over both space and time. 

\begin{figure}[tbp]
\centering
\begin{tabular}{c c} \includegraphics[width=0.47\textwidth]{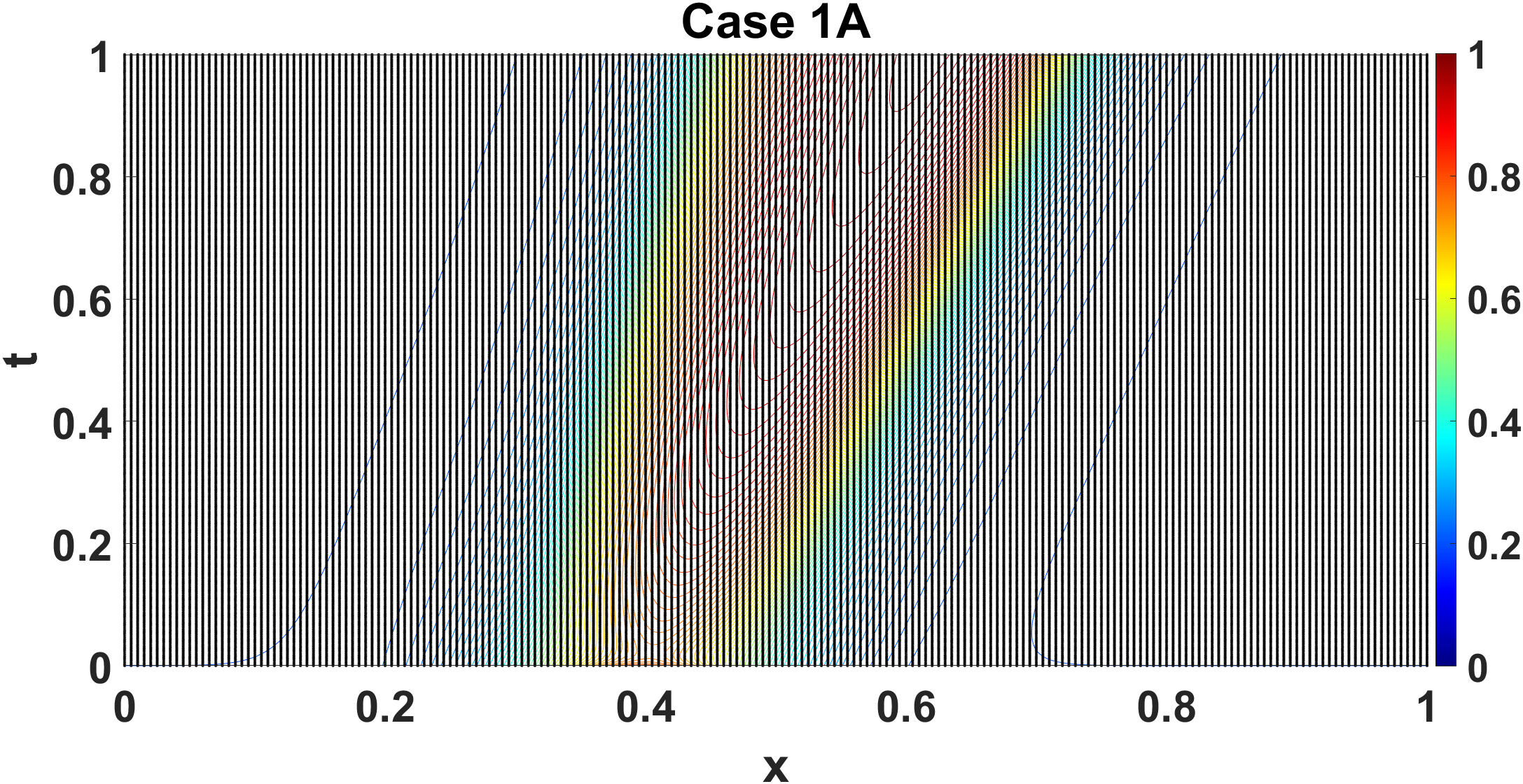} &
\includegraphics[width=0.47\textwidth]{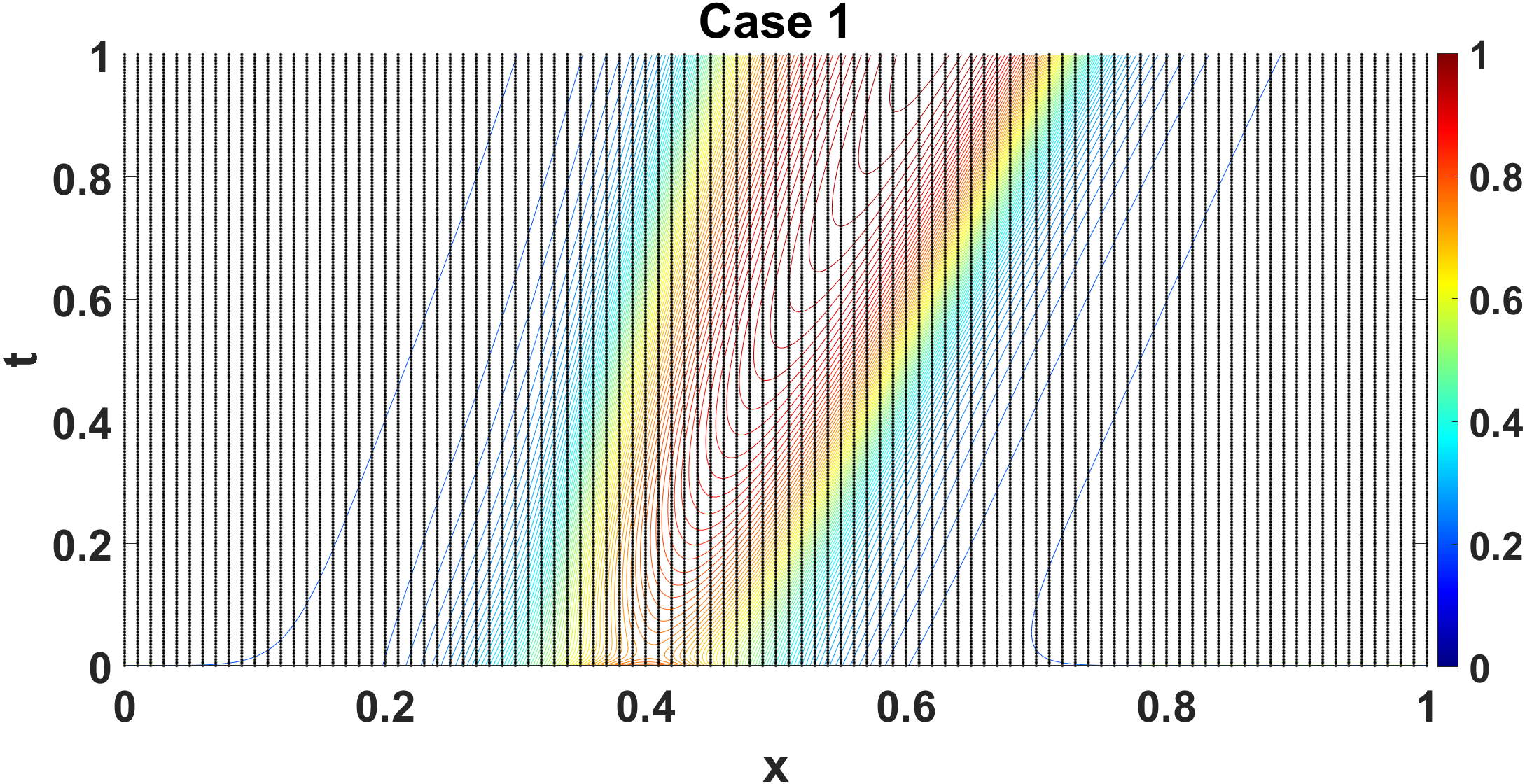} \\
\includegraphics[width=0.47\textwidth]{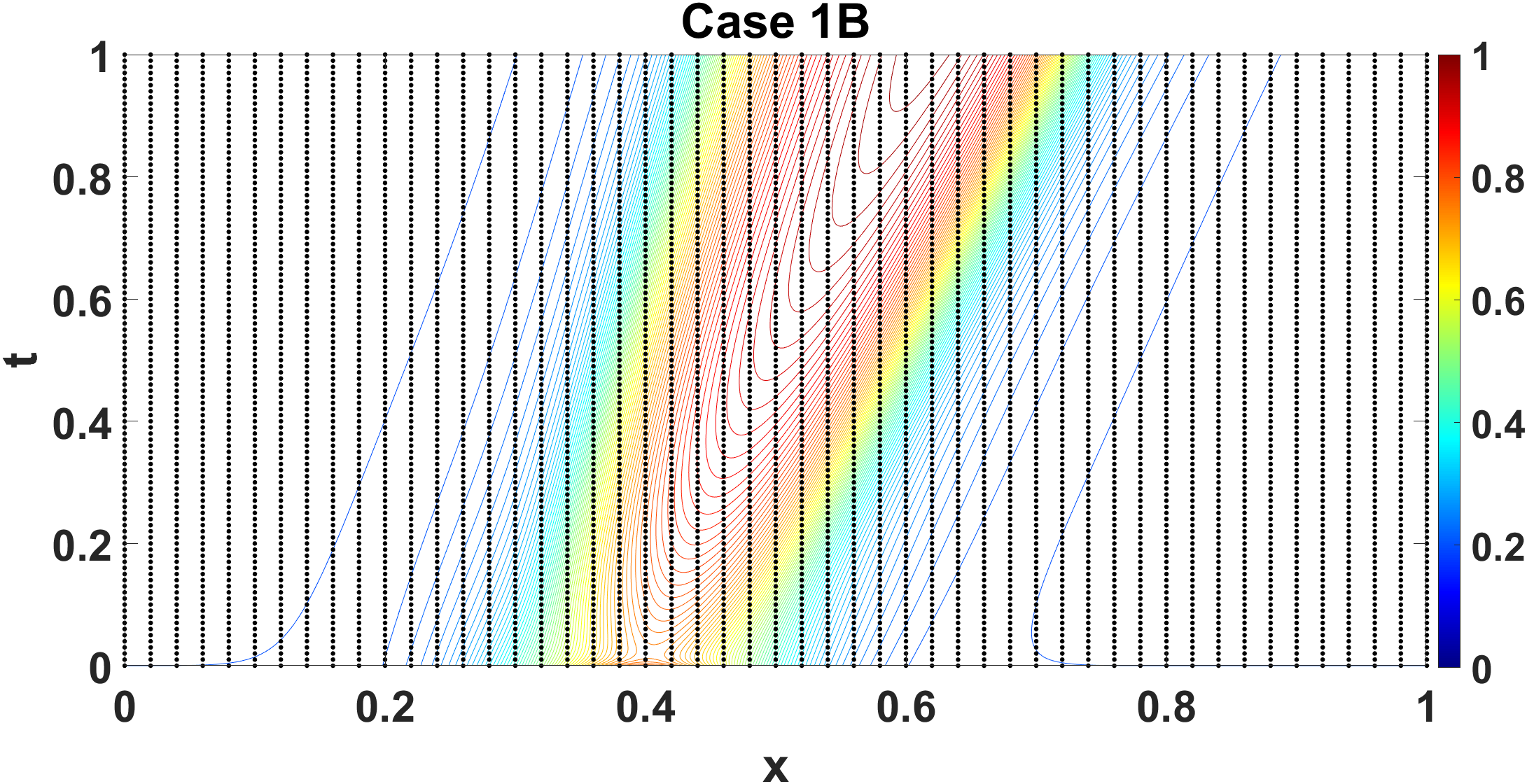} &
\includegraphics[width=0.47\textwidth]{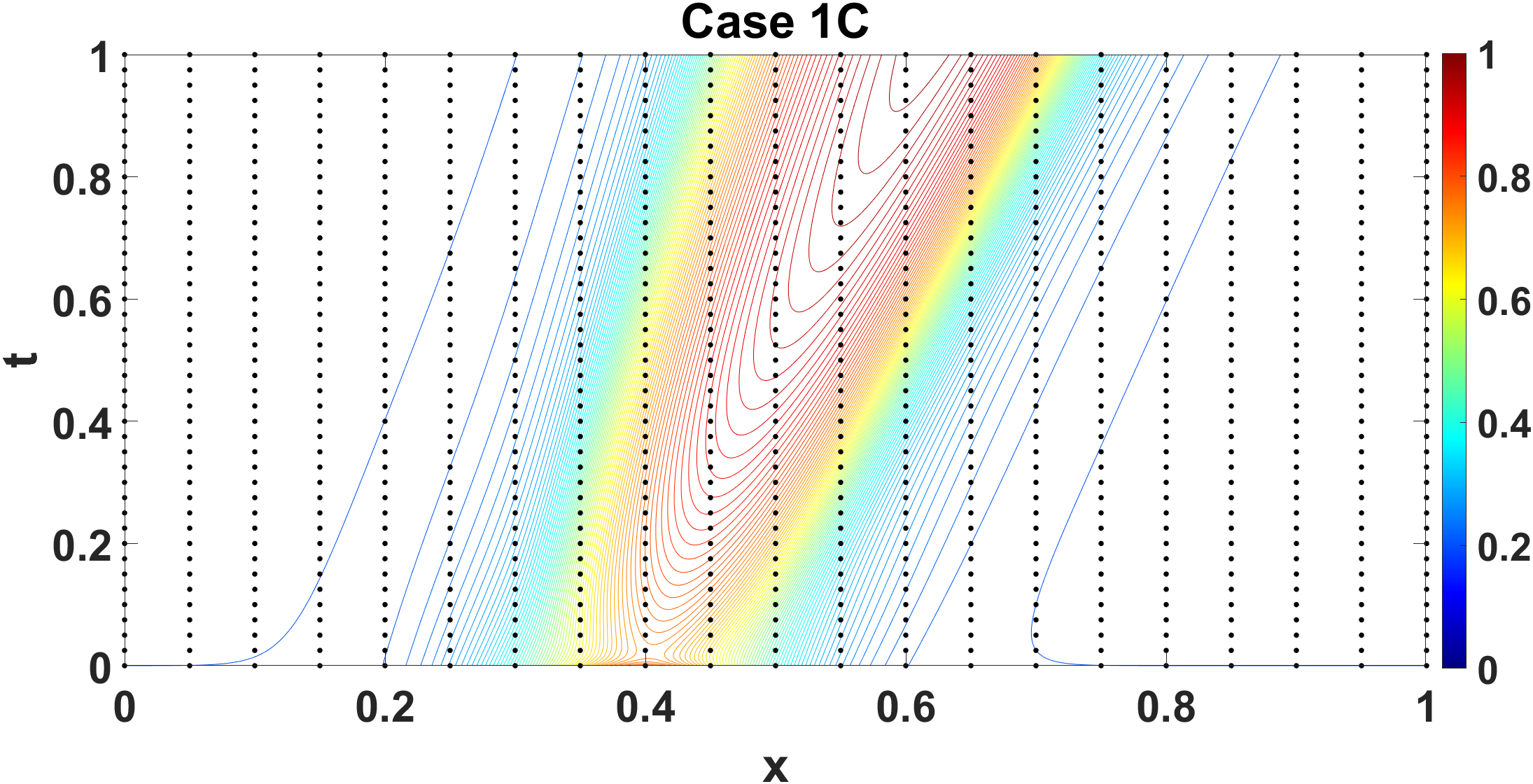}
\end{tabular}
\caption{Representation of the training dataset distributions across the temperature profile in both space and time for Case 1 (optimal), 1A, 1B, and 1C. Dot symbols represent the sampling data points used for training in each case.}
\label{fig:Sampling_points}
\end{figure}

While deep neural networks have resulted in a notable improvement in the spectrum of data-assimilation techniques, questions persist regarding their generalization capabilities. Particularly, concerns arise from the optimal configuration of the network architecture and the necessary number of training data points \citep{Yu2022}. Seeking to address these questions, training is conducted with datasets of different magnitudes (more or less data points from the optimal selection). Specifically, Case 1 (previously described example) includes 20.301 data points, while three new cases, 1A, 1B, and 1C, encompass 80.601, 5.151, and 861 data points for each state variable, respectively. The spatial and temporal sampling frequencies for generating these new training datasets are accordingly: $d_x = 0.5$ m, $d_t = 0.5$ s (for Case 1A), $d_x = 2$ m, $d_t = 2$ s (for Case 1B), and $d_x = 5$ m, $d_t = 5$ s (for Case 1C). \Cref{fig:Sampling_points} demonstrates the sampling distributions of the temperature training datasets between all the examined case studies over the entire spatiotemporal domain. 
Delving deeper into this, Table~\ref{table:Training_points} clearly illustrates that as the number of sampling points decreases, the accuracy of parameter estimation diminishes. However, even in cases with an insufficient number of data points (see Case 1B and Case 1C), ANN demonstrates promising capability in learning the unspecified parameters, despite experiencing significant oscillations during the training phase. On the other hand, while data can enhance model performance (see Case 1A), it also has the potential to exacerbate overfitting issues and poor generalization, meaning the model performs well on the training data but fails to predict accurately on unseen or new data. This issue is particularly challenging in real-world applications where data is often sparse and noisy, causing the model to prioritize fitting the training data perfectly at the expense of adhering to physical laws, which can result in unrealistic solutions. Techniques such as regularization, cross-validation, early stopping, and the incorporation of physics-based constraints can help mitigate the effects of overfitting. The convergence accuracy of Case 1 and Case 1A is ultimately equivalent, thus not requiring a greater amount of data to achieve results with enhanced precision (optimal trade-off between the number of data points and network complexity). In summary, depending on the nature of the problem and the NN's architecture, the magnitude (or "density") of the dataset can be considered one more mutable hyperparameter to tune. 

\begin{table}[tbp]
\centering
\caption{Learning capacity of PiNNs across multiple dataset sizes for Case 1 (optimal), Case 1A, Case 1B, and Case 1C.}
\label{table:Training_points}
\begin{tabular}{cccccccc}
\toprule
\textbf{Case} & \textbf{Data} & 
${d_x}\,  ([=]\, m)$ &
${d_t}\,  ([=]\, s)$ & \textbf{CPU Time}\, $\mathbf([=]\, h)$ &
$\hat{\textnormal{D}}([=]\,  m^2/s)$ & $\hat{\textnormal{u}}([=]\,  m/s)$ & $\hat{\textnormal{U}}([=]\,  W/m^2K)$ \\
\midrule
1A & 80.601 & 0.5 & 0.5 & 15 & 0.410 & 0.250 & 0.610\\
1 & 20.301 & 1 & 1 & 4 & 0.408 & 0.250 & 0.610\\
1B & 5.151 & 2 & 2 & 2 &   0.471 & 0.240 & 0.594\\
1C & 861 & 5 & 5 & 0.5 & 0.550 & 0.226 & 0.573\\
\bottomrule
\end{tabular}
\end{table}

\subsubsection{Parameter learning for 1D firefront with synthetic noisy data}
\label{sec:Case_2}

As previously mentioned, wildland fires are influenced by a myriad of consecutive biological and chemical processes, and continuous monitoring during an active fire event poses great challenges. Wildfires engender their own micro-climate characteristics (e.g., pyrocumulonimbus clouds, fire-induced winds, and pyrogenic circulations \citep{Goodrick2022}), introducing randomness into the path of the flames. Measurements are inherently subject to noise due to the instantaneous spatial and temporal variations in both climate conditions (e.g., precipitation rate, wind flow, and moisture content) and thermal degradation processes (e.g., flaming or smoldering-glowing oxidation). While current state-of-the-art fire prediction tools offer numerous simulation software options, they often lack the ability to provide data-driven estimations within the framework of model calibration. Considering that noisy data is the general case rather than the exception, it is reasonable to investigate the robustness of PiNNs in learning the unknown wildfire spreading model parameters, even under these challenging conditions.

\begin{figure}[tbp]\centering
\begin{tabular}{cc} \includegraphics[width=0.47\textwidth]{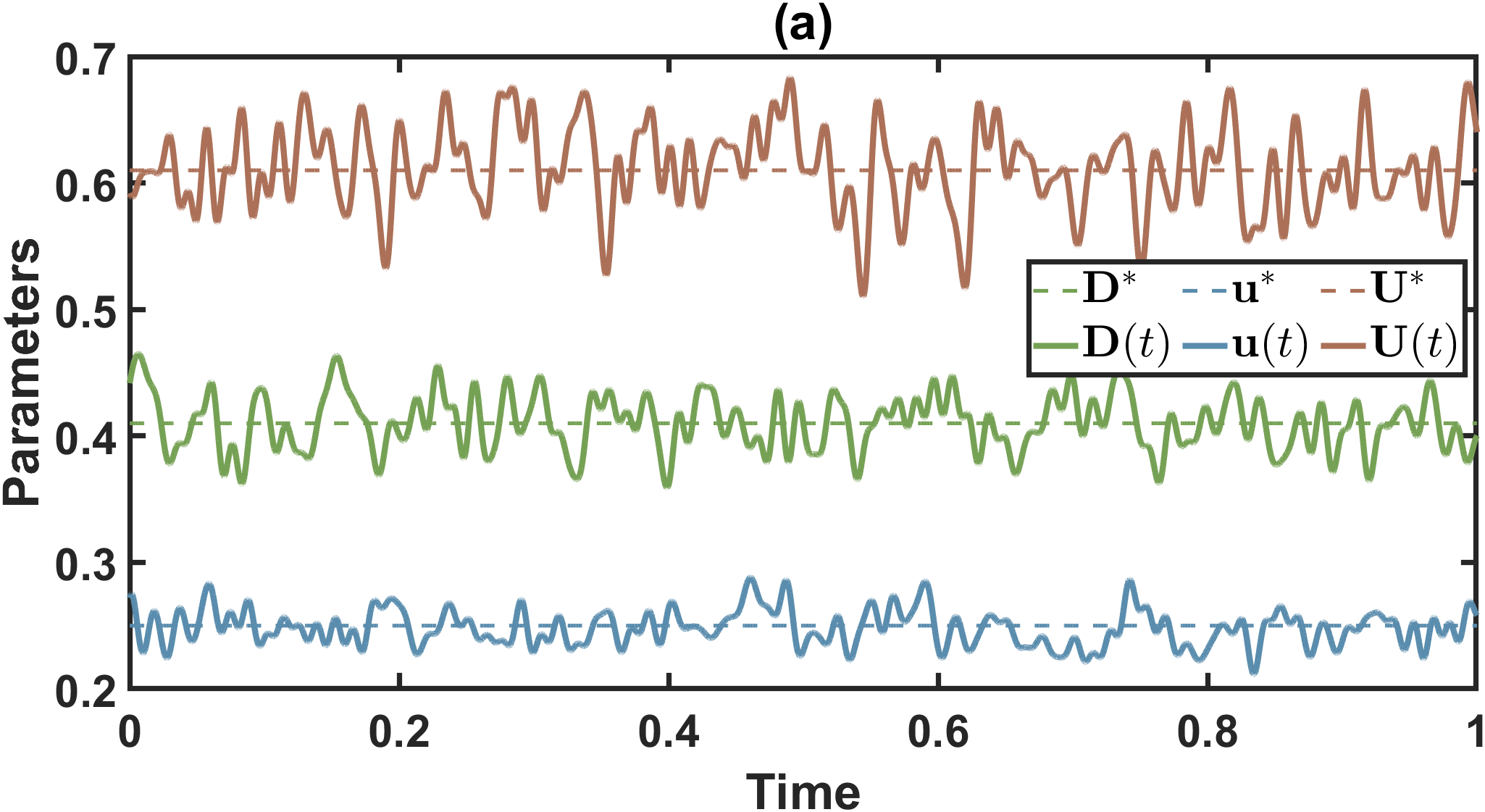} &
\includegraphics[width=0.46\textwidth]{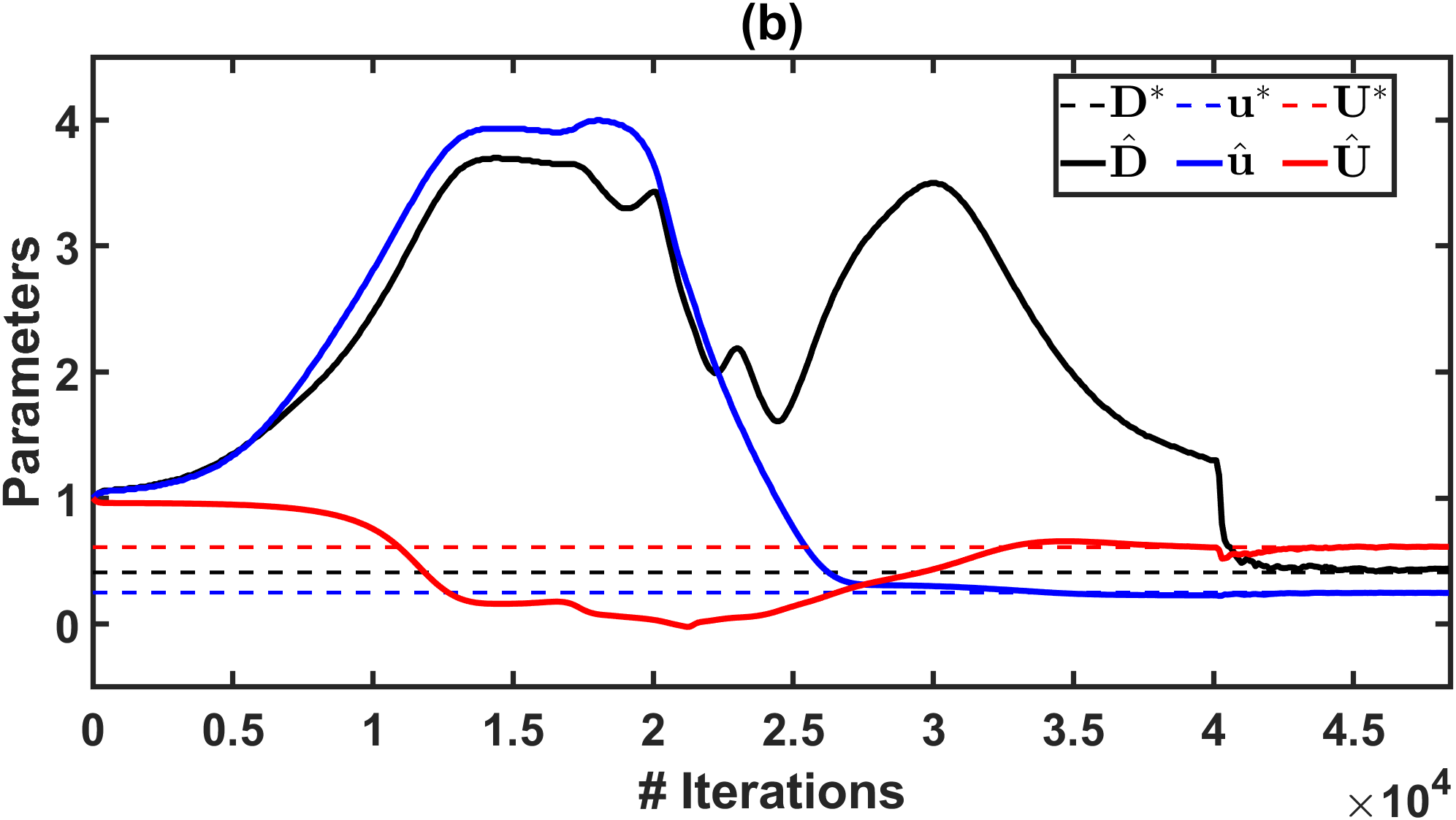}
\end{tabular}
\caption{(a) Dotted lines correspond to the constant parameter vector, $\bm{\theta}^{*} = \left[0.41\,\, 0.25\,\, 0.61\right]^\T$, used for generating the training dataset for \emph{Case Study 1}. Solid lines correspond to the temporally perturbed parameter vector, $\bm{\theta}(t) = [\textnormal{D}(t)\,\, \textnormal{u}(t)\,\,\textnormal{U}(t)]^\T$, employed for generating the training dataset for \emph{Case Study 2}. (b) Parameter learning and convergence process. Predicted vector of the three 
 model parameters, $\hat{\bm{\theta}} = [0.44\,\, 0.242\,\, 0.613]^\T$. True vector for generating the training dataset, $\bm{\theta}^{*} = \left[0.41\,\, 0.25\,\, 0.61\right]^\T$.}
\label{fig:Noisy_1D_Data}
\end{figure}

In \emph{Case Study 2}, the formulation and assumptions remain identical to those of \emph{Case Study 1}, with the only exception being the incorporation of simulated training data prepared by introducing noise (model error). Specifically, it is assumed that the parameters, 
$\bm{\theta}(t) = [\textnormal{D}(t)\,\, \textnormal{u}(t)\,\,\textnormal{U}(t)]^\T = [\textnormal{D}^{*}+\Delta \textnormal{D}(t)\,\, \textnormal{u}^{*}+\Delta \textnormal{u}(t)\,\,\textnormal{U}^{*}+\Delta \textnormal{U}(t)]^\T$ $\forall \, t \in \Upsilon$, 
fluctuate in time around their nominal values, $\bm{\theta}^{*} = [\textnormal{D}^{*}\,\, \textnormal{u}^{*}\,\,\textnormal{U}^{*}]^\T = \left[0.41\,\, 0.25\,\, 0.61\right]^\T$. Gaussian stochastic processes are used to model these fluctuations, 
$\Delta \textnormal{D}(t),\,\, \Delta \textnormal{u}(t),\,\, \Delta \textnormal{U}(t)$,  
with a zero mean and a correlation structure given by $\delta_{\varphi}^2 e^{-\frac{|t_j-t_i|}{\zeta_{\varphi}}}$, quantifying the correlation in time. In this context, $\delta_{\varphi}^2$ denotes the magnitude of the fluctuations, $\zeta_{\varphi}$ represents the temporal correlation time, and the subscript $\varphi$ stands for either the $\textnormal{D}$, $\textnormal{u}$, or $\textnormal{U}$ parameter. Specifically, a model error of $\delta_{\varphi}=5\%$ and a correlation time of $\zeta_{\varphi}= 0.005$ is selected, signifying fluctuations around the nominal values, $\bm{\theta}^{*}$, that can reach up to 15$\%$. As our model formulation exhibits functionality at a length scale of hundreds of meters, without addressing the full-scale geophysical problem structure, temporal variations are dominant, thus avoiding capturing possible spatial perturbations. 
\Cref{fig:Noisy_1D_Data}(a) provides the comparison between the constant parameter values used in \emph{Case Study 1} and the fluctuated values employed in this example to generate the training datasets.

The outcomes are highly promising, with PiNNs facilitating the convergence of the unknown parameters to their nominal values, $\bm{\theta}^{*}$, absorbing the substantial fluctuations present in the training dataset. However, due to the appearance of external noise, the network explores a broader range of feasible parameter values, resulting in delayed total convergence, typically observed after 35.000 iterations. \Cref{fig:Noisy_1D_Data}(b) depicts the obtained results on the
learning trajectories of the unknown parameters, successfully converging to $\hat{\bm{\theta}} = [\hat{\textnormal{D}}\,\, \hat{\textnormal{u}}\,\,\hat{\textnormal{U}}]^\T = [0.44\,\, 0.242\,\, 0.613]^\T  \approx \bm{\theta}^{*}$. This resilience exemplifies a significant benefit of PiNNs in handling noisy data (as observed in real-world measurements) \citep{Lu2021} compared to traditional data-assimilation methods with restricted tolerance to noise.

\subsection{PiNN for 2D firefront}
\label{sec:2D_modeling}

The preceding case studies highlight the importance of PiNNs in the case of 1D wildfire propagation for simplifying complex dynamics, enabling faster computation, and easier analysis of fundamental spread mechanisms \citep{Finney2022}. While sophisticated three-dimensional (3D) simulation software includes both multi-phase combustion models (operating at centimeter scales) \citep{Morvan2004}, wildfire simulation models (operating at meter scales) \citep{Linn2005}, and atmospheric boundary layer models (operating at kilometer scales) \citep{Coen2013}, their practical applicability is limited by the considerable theoretical expertise and high-performance computing resources needed. Consequently, these software solutions are confined only to research settings. In contrast, the 2D modeling of the firefront presented herein serves as the solution for a time-constrained, physics-and-data-informed digital tool capable of incorporating measurements from both ground and airborne sensing platforms (e.g., IR imager data). This tool can instantly inform decision-making authorities to facilitate coordination and aggressive suppression tactics, such as evacuation paths, reforestation policies, and fuel breaks, to mitigate wildfire spread, particularly during emergency situations where firefighting efforts risk being overwhelmed.

\begin{figure}[tbp]\centering
\begin{tabular}{c c c} \includegraphics[width=0.348\textwidth]{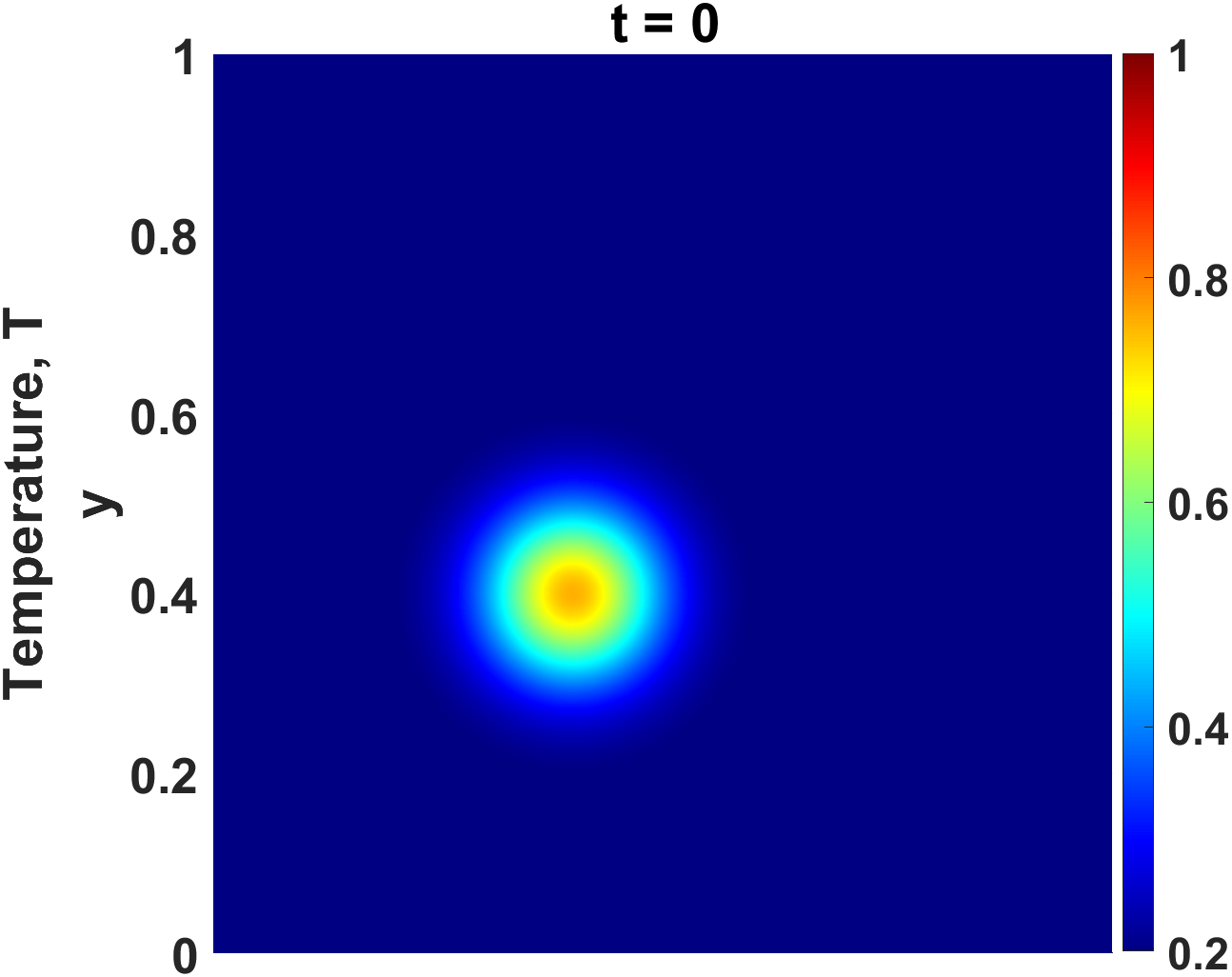} &
\includegraphics[width=0.285\textwidth]{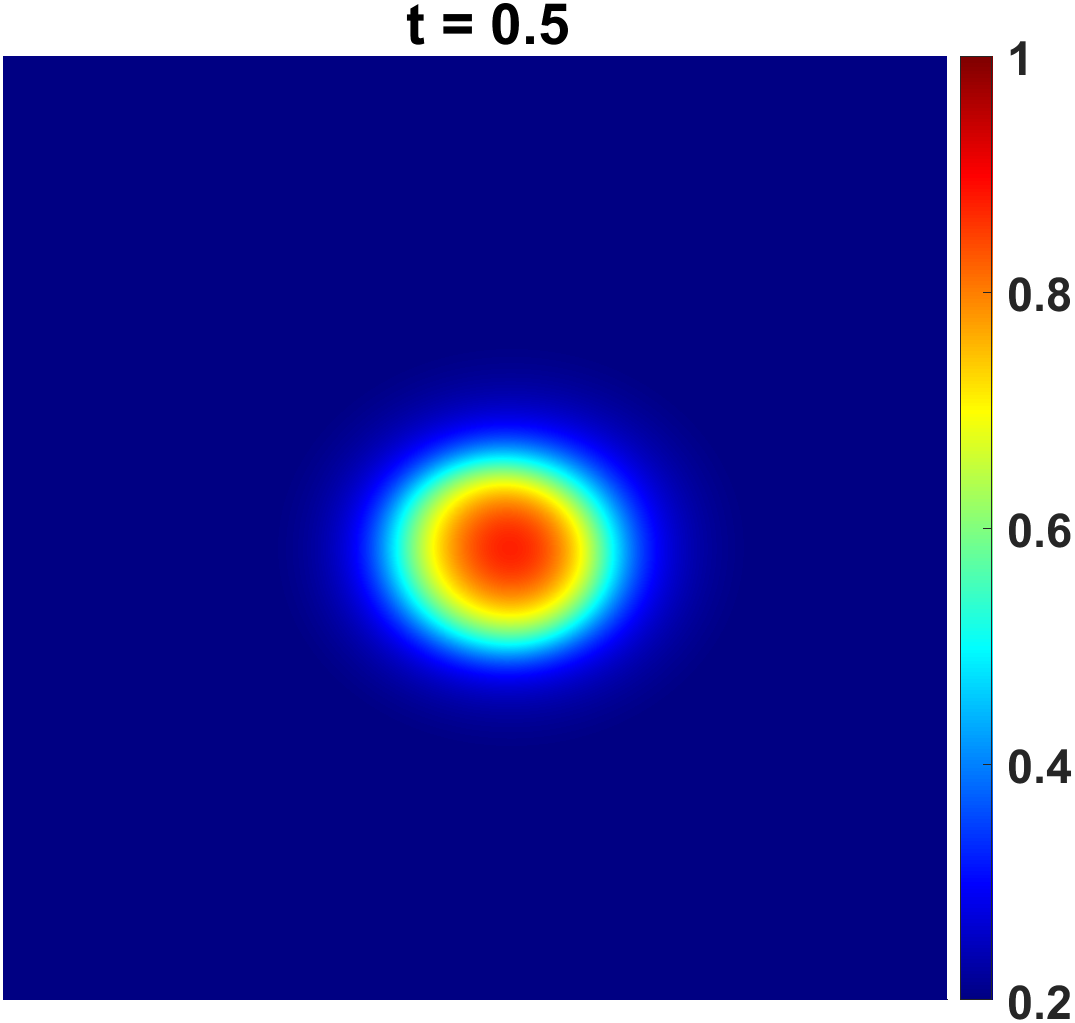} &
\includegraphics[width=0.285\textwidth]{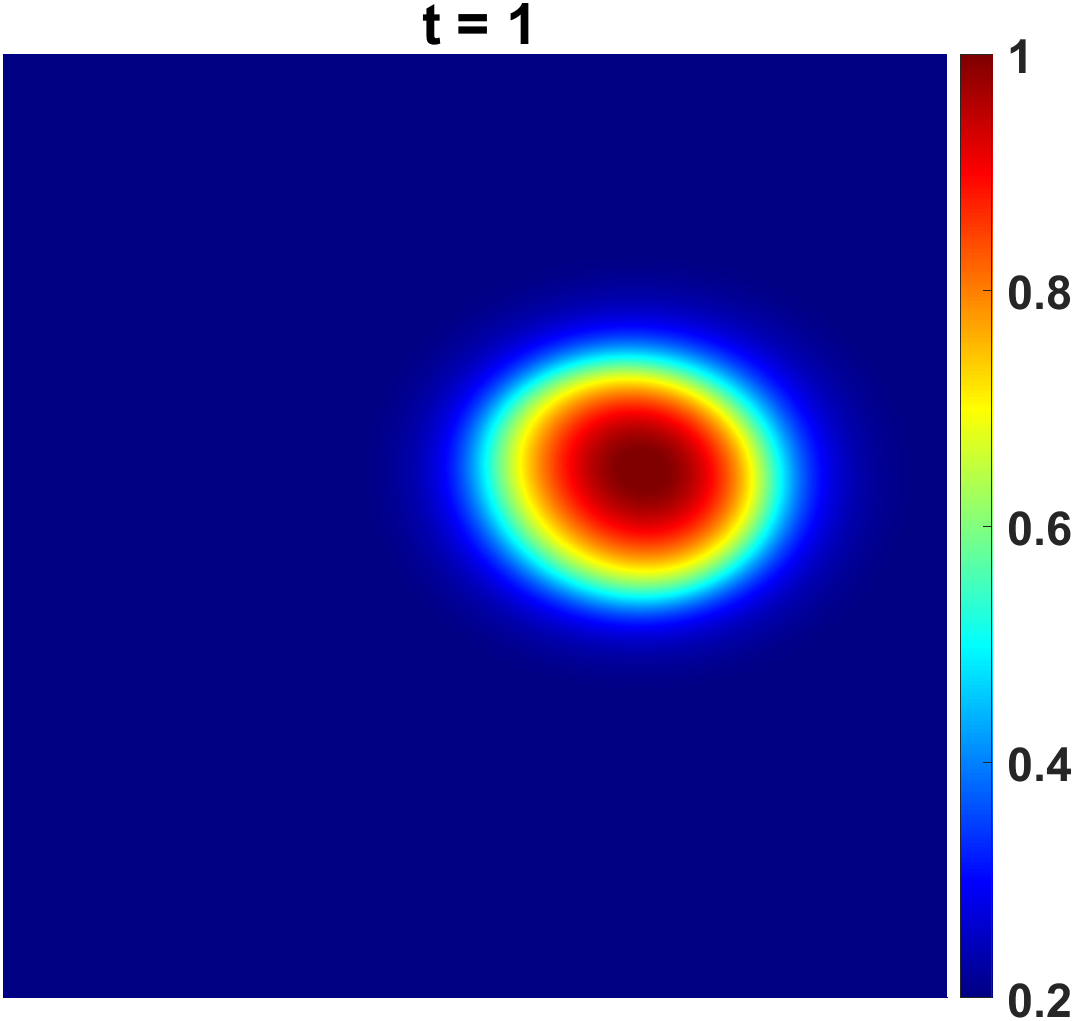}  \\
\includegraphics[width=0.35\textwidth]{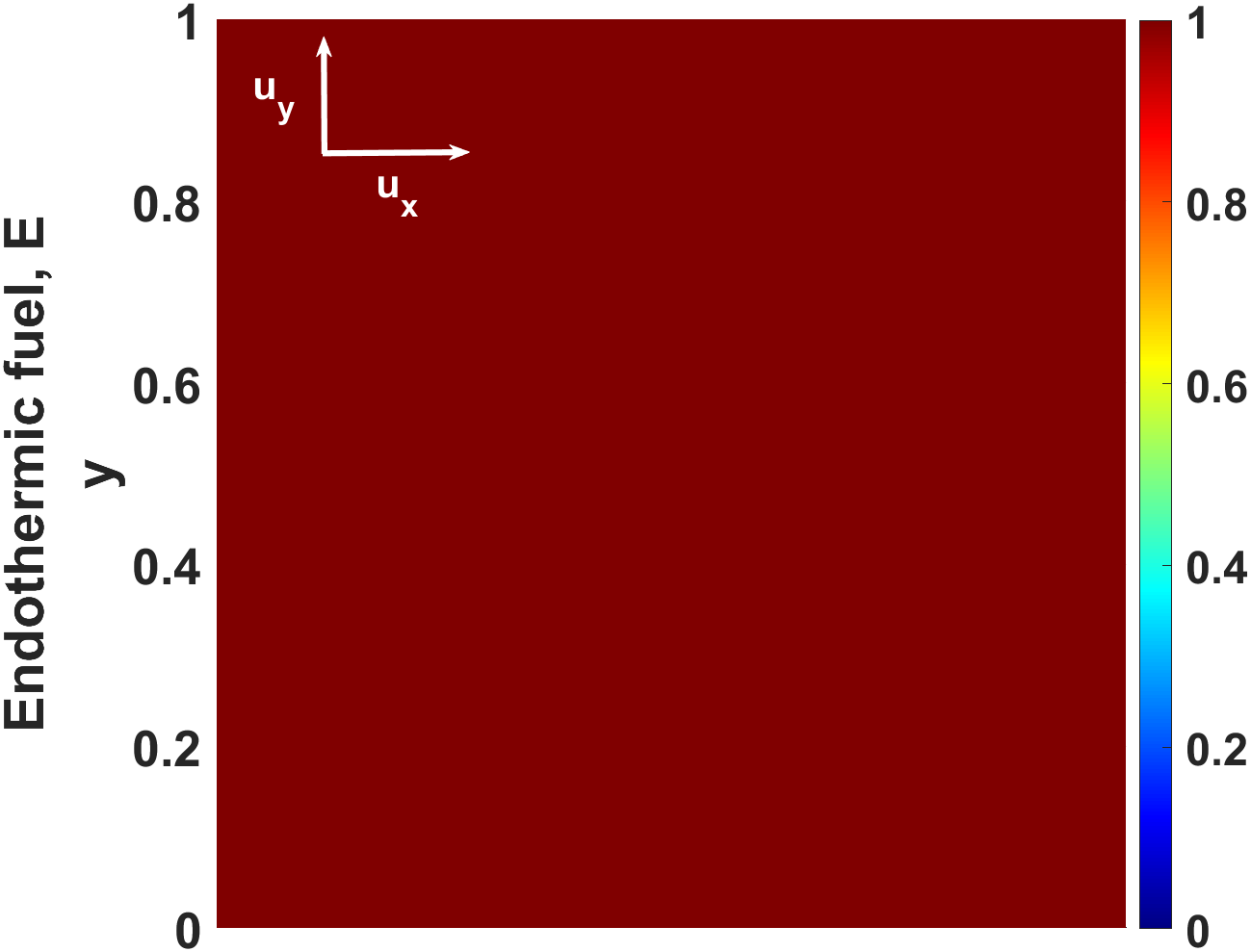} &
\includegraphics[width=0.285\textwidth]{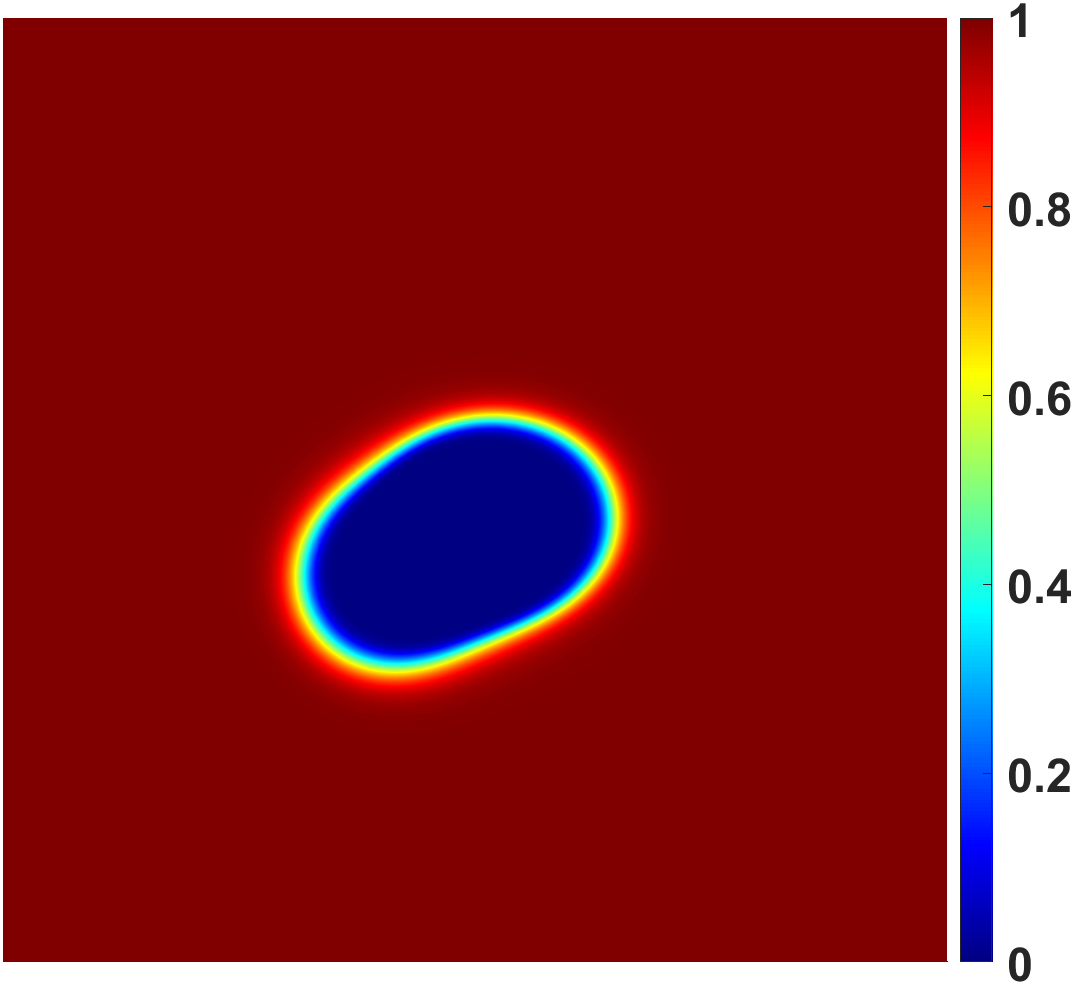} &
\includegraphics[width=0.285\textwidth]{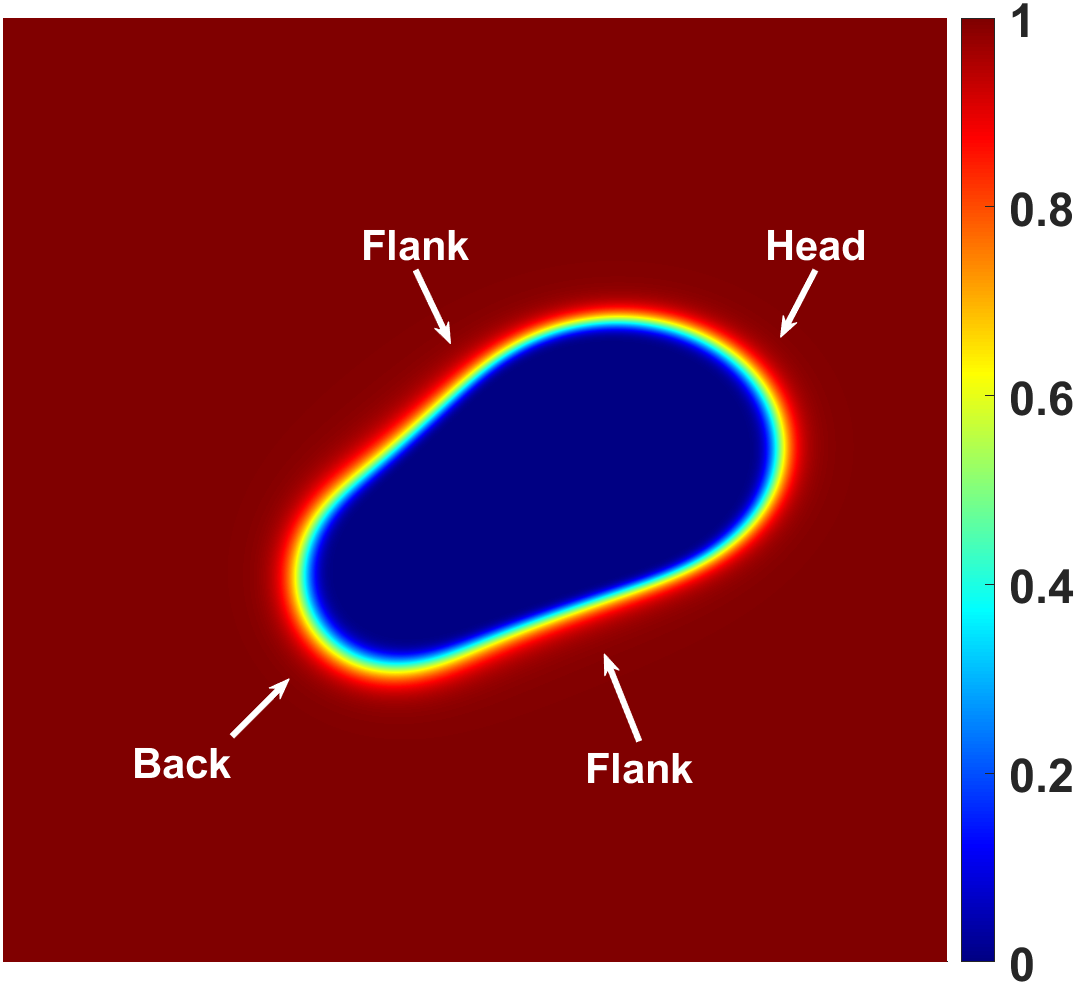}  \\ 
\includegraphics[width=0.35\textwidth]{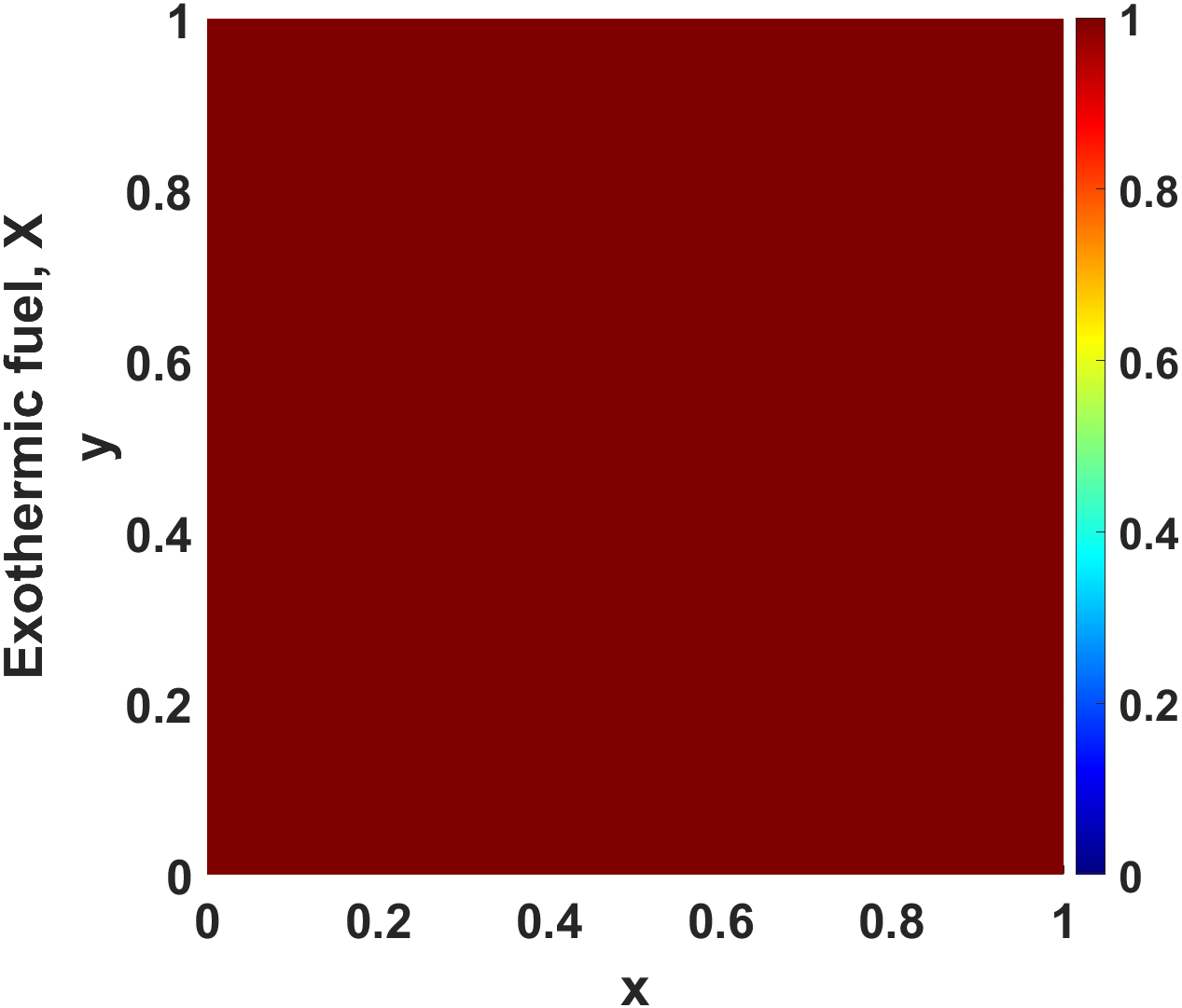} &
\includegraphics[width=0.289\textwidth]{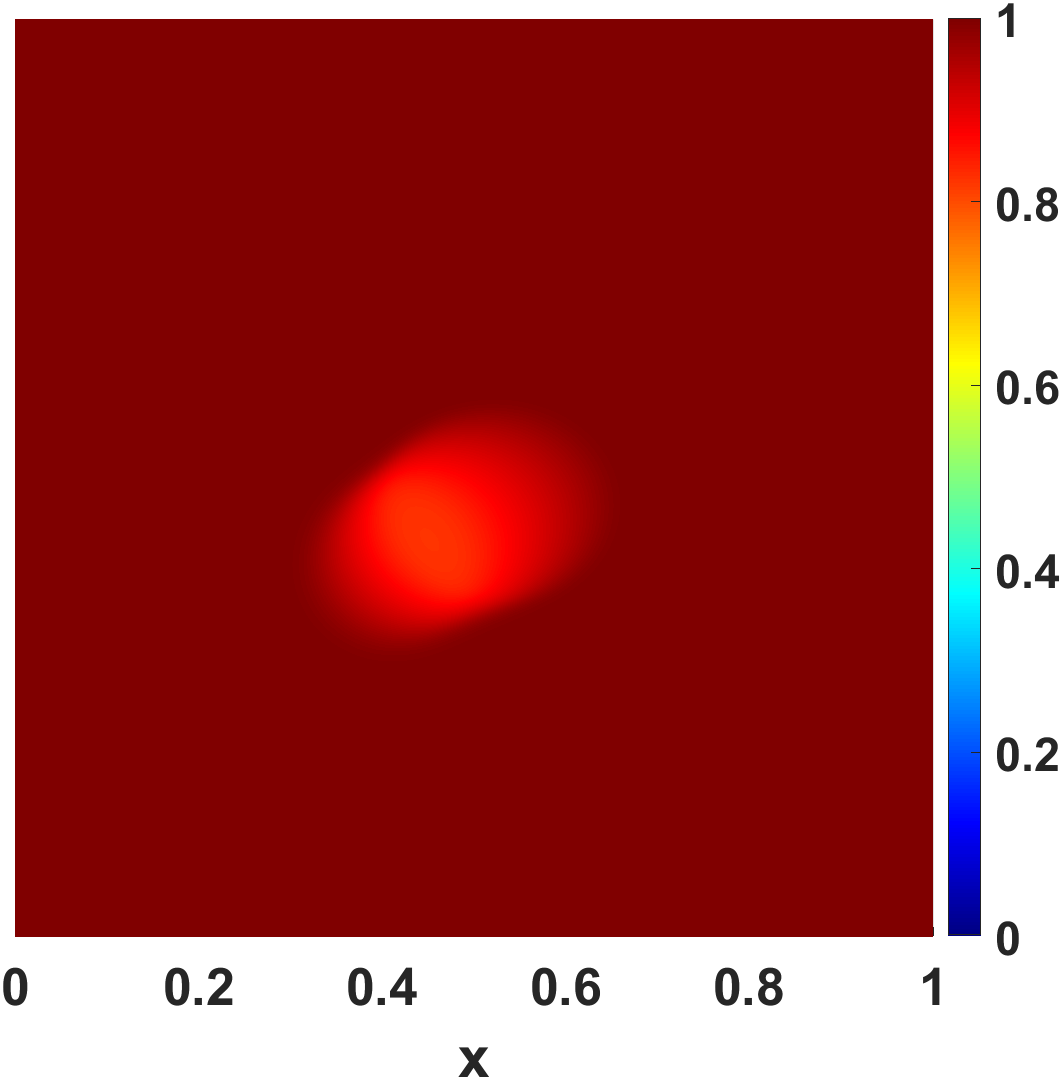} &
\includegraphics[width=0.289\textwidth]{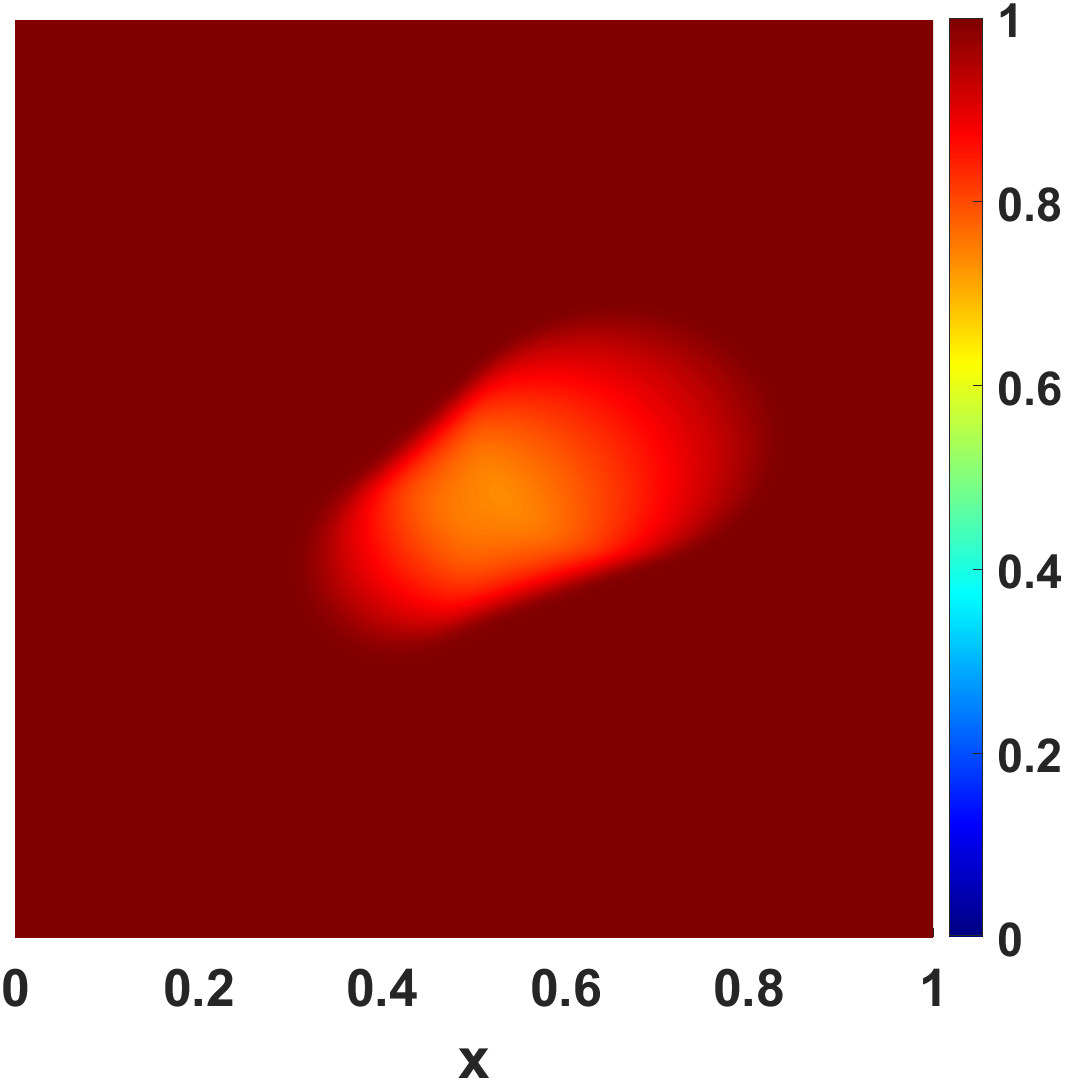} 
\end{tabular}
\caption{Prototype simulation for the 2D spatiotemporal firefront
generated by the wildfire spreading model. Each row represents the state profile of temperature ($T$), endothermic fuel ($E$), and exothermic fuel ($X$). The columns denote the three time instants, $t = 0, 0.5$, and $1$. The results are provided in a dimensionless form.}
\label{fig:2D_simulation}
\end{figure}

In both \emph{Case Study 3} (see \Cref{sec:Case_3}) and \emph{Case Study 4} (see \Cref{sec:Case_4}), the focus lies on the precise estimation of the unspecified parameters within the 2D wildfire spreading model. The formulation mirrors the assumptions employed in the 1D scenario, ensuring consistency with the fundamental physics involved. Identical initial and boundary conditions persist (yet extended in two dimensions), considering that boundaries remain unaffected by fire progression. \Cref{fig:2D_simulation} depicts the prototype simulation used to generate the training sample utilized during the execution of the learning algorithm. Each row in the figure represents the evolution of temperature ($T$), endothermic fuel ($E$), and exothermic fuel ($X$), while columns correspond to three specific non-dimensional time instants, $t = 0, 0.5$, and $1$, respectively, captured during the simulation. The depicted figures validate the state-of-the-art firefront geometry, resampling a parabolic shape (horseshoe geometry associated with free-burning fires) characterized by increased forward and intermediate lateral expansion alongside decelerated backward flow \citep{Vogiatzoglou2024}.

As for the selected neural network architecture, it is similar to that of the 1D modeling formulation, with the notation set as follows: $L$ = 5, $N_{0}$ = 3, $N_{\ell}$ = 20 for $\ell =$ 1, 2, 3, 4, and
$N_{5}$ = 3. Once again, we performed an independent analysis to identify the optimal neural network architecture. Notably, this small-scale network offers a surprising advantage in our formulation, demonstrating the effectiveness of the proposed approach even with a simplified structure. Specifically, the network includes one (1) input layer with three (3) neurons $\in \R^3$ (representing spatiotemporal coordinates $x, y \in \Omega$ and $t \in \Upsilon$), four (4) hidden layers with twenty (20) neurons each, and one (1) output layer with three (3) neurons $\in \R^3$, corresponding to the predicted state variables, $\hat{{\cal P}} = [\hat{T} \  \hat{E} \ \hat{X}]^\T$ (same output quantities of interest). Therefore, the parameterized set notation for this network example reads $\mathbf{A} = \left\{\W^{1}, \W^{2}, \W^{3},\W^{4}, \W^{5}, \b^{1}, \b^{2}, \b^{3}, \b^{4}, \b^{5}\right\} \in \R^{1403}$. Regarding the activation function and optimization algorithms employed, a logistic sigmoid function as well as the Adam algorithm with a learning rate equal to 0.0003 are initially applied for 60.000 iterations, followed by the L-BFGS optimizer to expedite convergence.

\subsubsection{Parameter learning for 2D firefront with synthetic  data}
\label{sec:Case_3}
 
Considering the problem in two dimensions, the diffusion coefficient, $\mathbf{D}$, and the mean gaseous velocity, $\mathbf{u}$, necessitate two components each in the $x-$ and $y-$directions, resulting in a total of five model parameters for learning. Once again, synthetic data is generated using the same discretization numerical methods and tested with known predefined values, $\bm{\theta}^{*} = [\textnormal{D}_x^{*}\,\, \textnormal{D}_y^{*}\,\, \textnormal{u}_x^{*}\,\, \textnormal{u}_y^{*}\,\, \textnormal{U}^{*}]^\T = \left[0.74\,\, 0.41\,\, 0.35\,\, 0.2\,\, 0.4\right]^\T$. The simulation encompasses a rectangular spatial domain of $x \in \left[0, 100\right]$ m (streamwise direction), $y \in \left[0, 100\right]$ m (spanwise direction), and a time interval of $t \in \left[0, 200\right]$ s, while the training dataset comprises simulated points sampled at uniform intervals of $d_x =$ 4 m, $d_y =$ 4 m in space, and $d_t =$ 4 s in time, contributing to a total of 34.476 training points for each state variable. This dataset size has been found to be optimal, striking a compromise between having sufficient points for training and minimizing architecture complexity.

\begin{figure}[tbp]
\centering
\includegraphics[width=.55\textwidth]{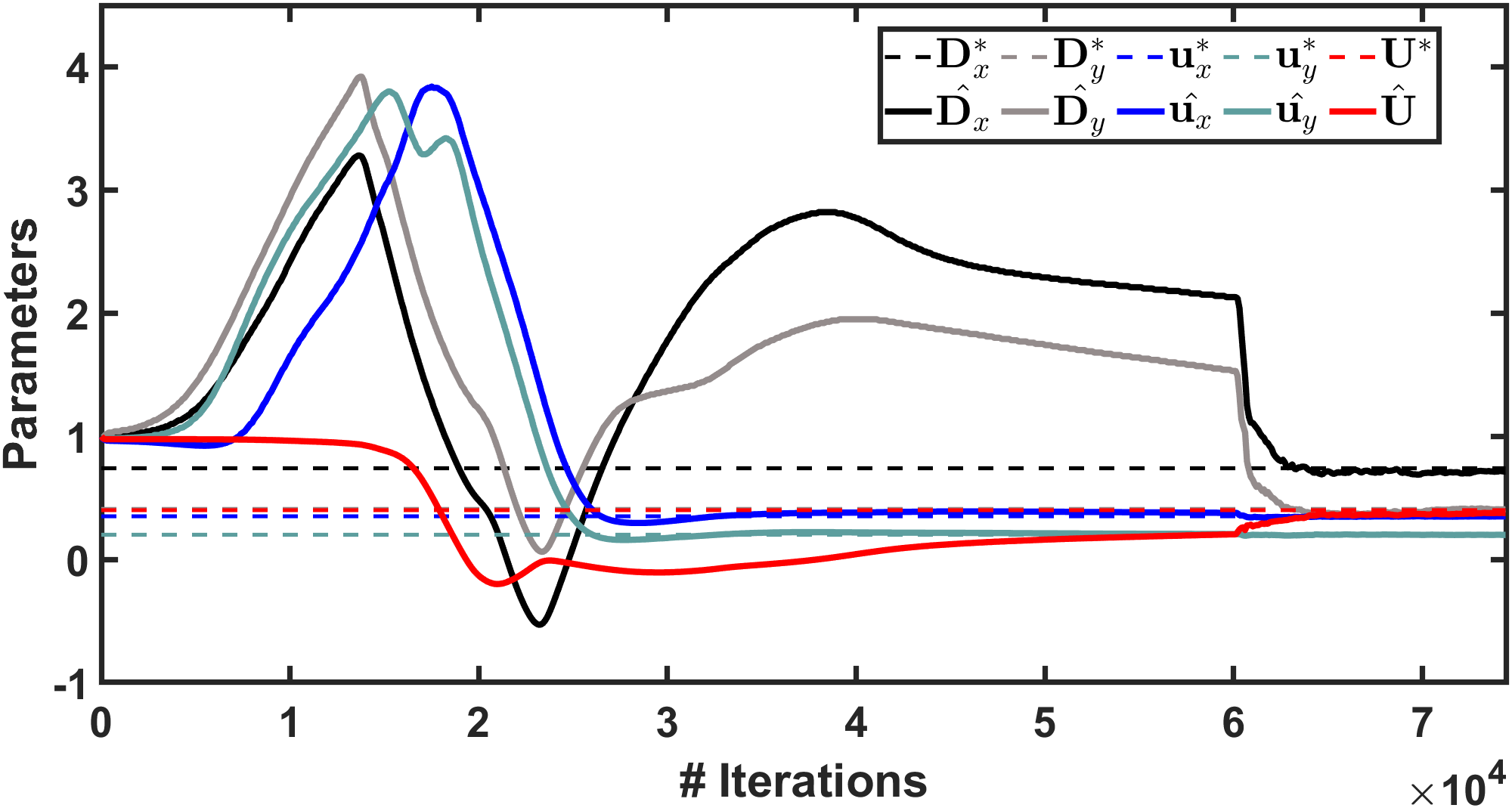}
\caption{Parameter learning and convergence in the 2D firefront of the wildfire spreading model. The predicted vector of the five unknown model parameters is $\hat{\bm{\theta}} = [0.724\,\, 0.41\,\, 0.35\,\,0.2\,\, 0.382]^\T$, while the true vector utilized for generating the training dataset is $\bm{\theta}^{*} = [0.74\,\, 0.41\,\, 0.35\,\,0.2\,\, 0.4]^\T$.}
\label{fig:Parameter_learning_2d}
\end{figure}

By initializing all unknown model parameters to the predefined value 1.0 (as an initial guess), PiNNs exhibit rapid convergence towards their nominal values, $\bm{\theta}^{*}$, upon completion of the total number of training iterations (11 h of CPU time). Specifically, as illustrated in \Cref{fig:Parameter_learning_2d}, certain model parameters ($\textnormal{u}_x, \textnormal{u}_y$) reached their nominal values before 30.000 iterations, while the remaining parameters ($\textnormal{D}_x, \textnormal{D}_y, \textnormal{U}$) required additional steps to achieve complete alignment. The L-BFGS algorithm, taking advantage of the Hessian matrix, accelerated the convergence rate (particularly for the three sensitive parameters) following the completion of the Adam algorithm. This enhancement contributed to the final prediction, $\hat{\bm{\theta}} = [\hat{\textnormal{D}}_x\,\, \hat{\textnormal{D}}_y\,\, \hat{\textnormal{u}}_x\,\, \hat{\textnormal{u}}_y\,\, \hat{\textnormal{U}}]^\T = [0.724\,\, 0.41\,\, 0.35\,\,0.2\,\, 0.382]^\T \approx \bm{\theta}^{*}$. PiNNs explored the feasible parameter range and facilitated the transition of specific parameters from negative to positive values.

\begin{figure}[tbp]\centering
\begin{tabular}{cc} \includegraphics[width=0.45\textwidth]{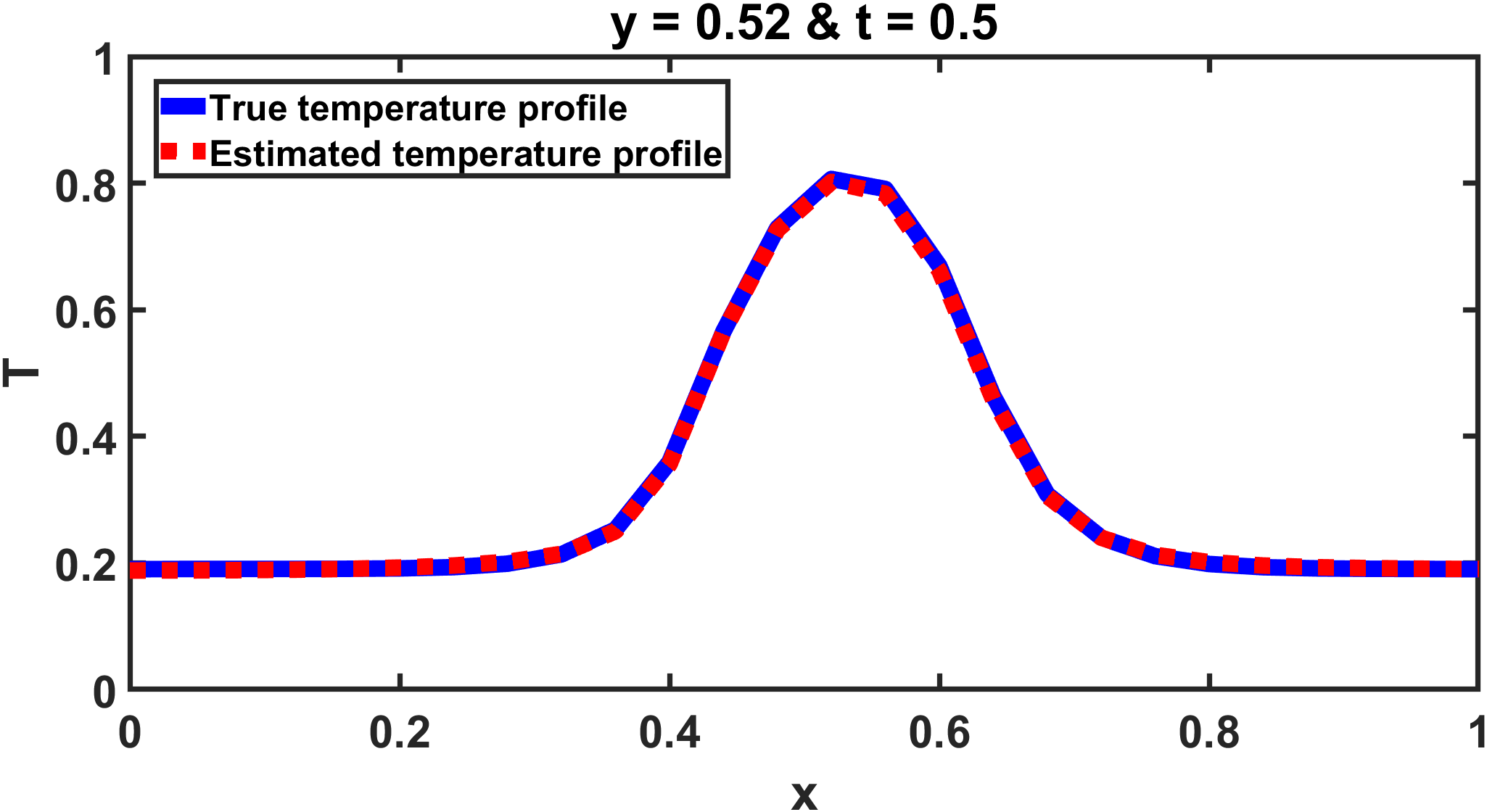} &
\includegraphics[width=0.45\textwidth]{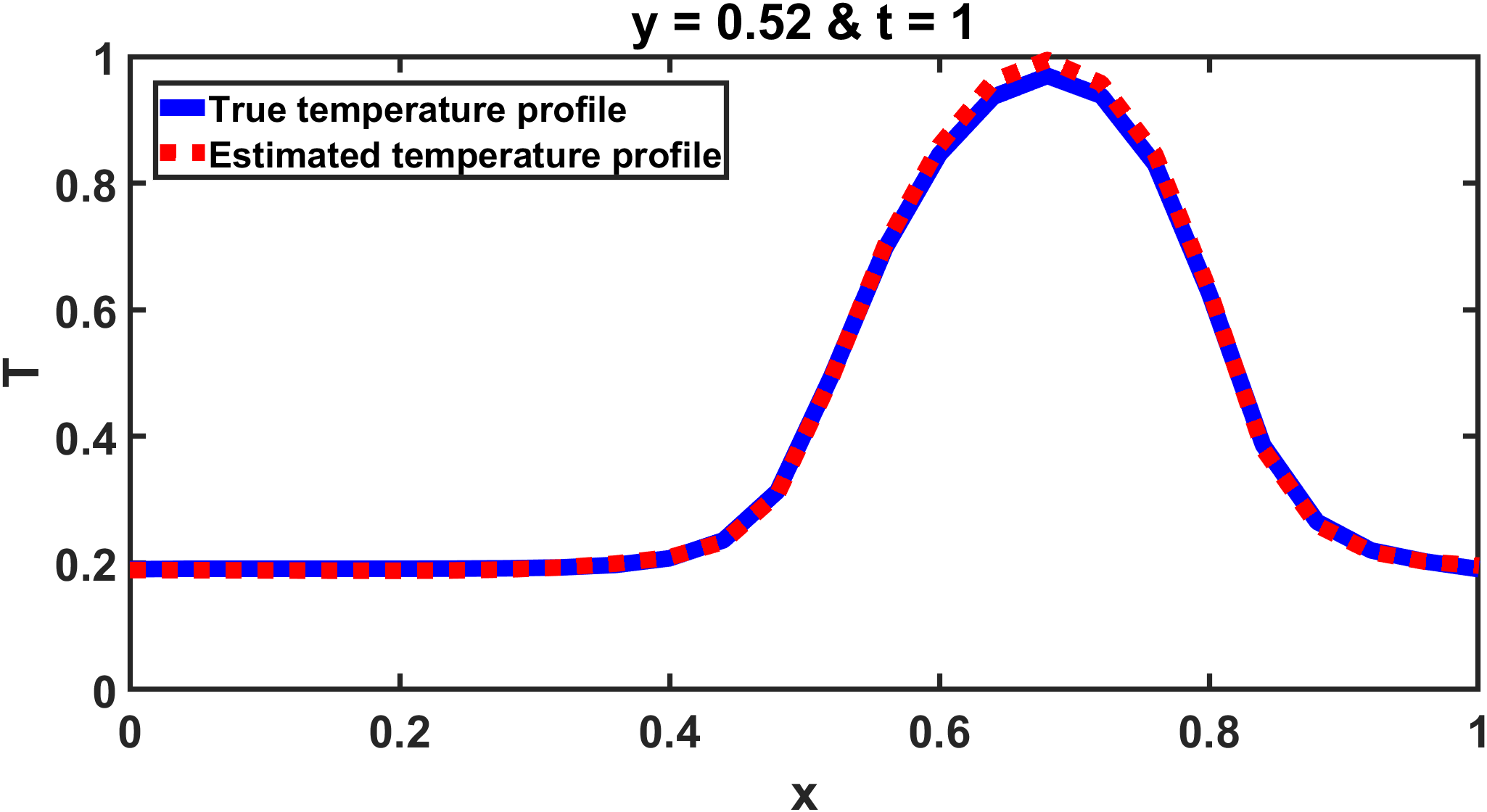}
\end{tabular}
\caption{Comparison of the temperature evolution between the explicit solution and PiNNs prediction at two non-dimensional time instants, $t = 0.5$, and $t = 1$, for all nodes across the $x-$direction with $y =$ 0.52.}
\label{fig:2D_time_comparison}
\end{figure}

Finally, the outcomes demonstrate enhanced accuracy, being evident by comparing the temperature distribution between the predictions generated by PiNNs and the explicit solution derived from the mathematical system. In particular, \Cref{fig:2D_time_comparison} illustrates the temperature profile across the entire domain, examining the differences between the analytical and estimated solution for all nodes in the $x$-direction. This evaluation is performed for nodes with a fixed $y$-coordinate equal to 0.52 and for two distinct time instants, $t =$ 0.5 and 1. The curves are almost identical, signifying a similar spatiotemporal evolution of the firefront. All the consistent robustness across various PiNN operations carries significant implications, as it enables the tackling of more intricate problems across a broader range of dimensions.

\subsubsection{Parameter learning for 2D firefront with synthetic noisy data}
\label{sec:Case_4}

\begin{figure}[tbp]
\centering
\begin{tabular}{cc} \includegraphics[width=0.47\textwidth]{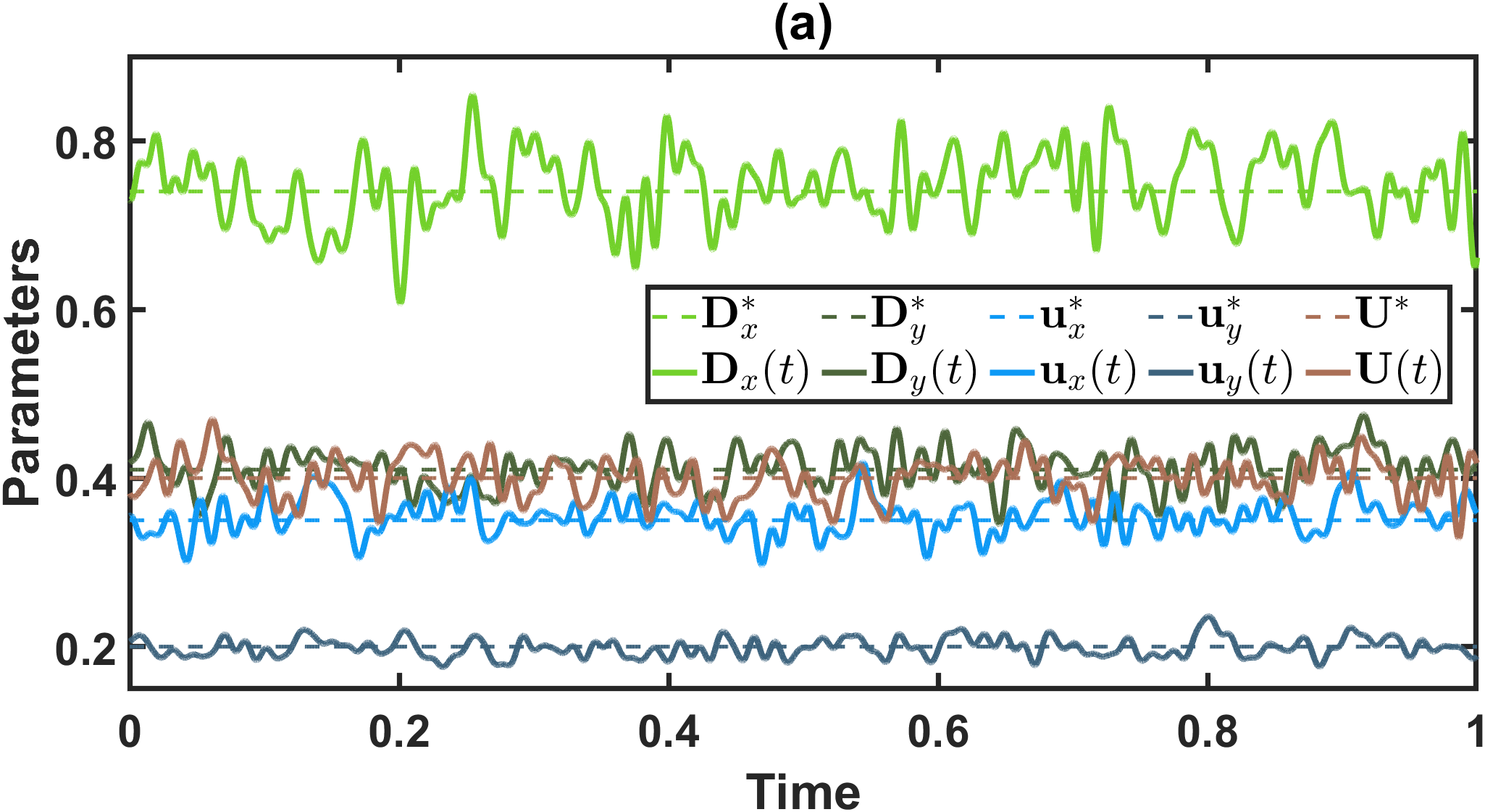} &
\includegraphics[width=0.47\textwidth]{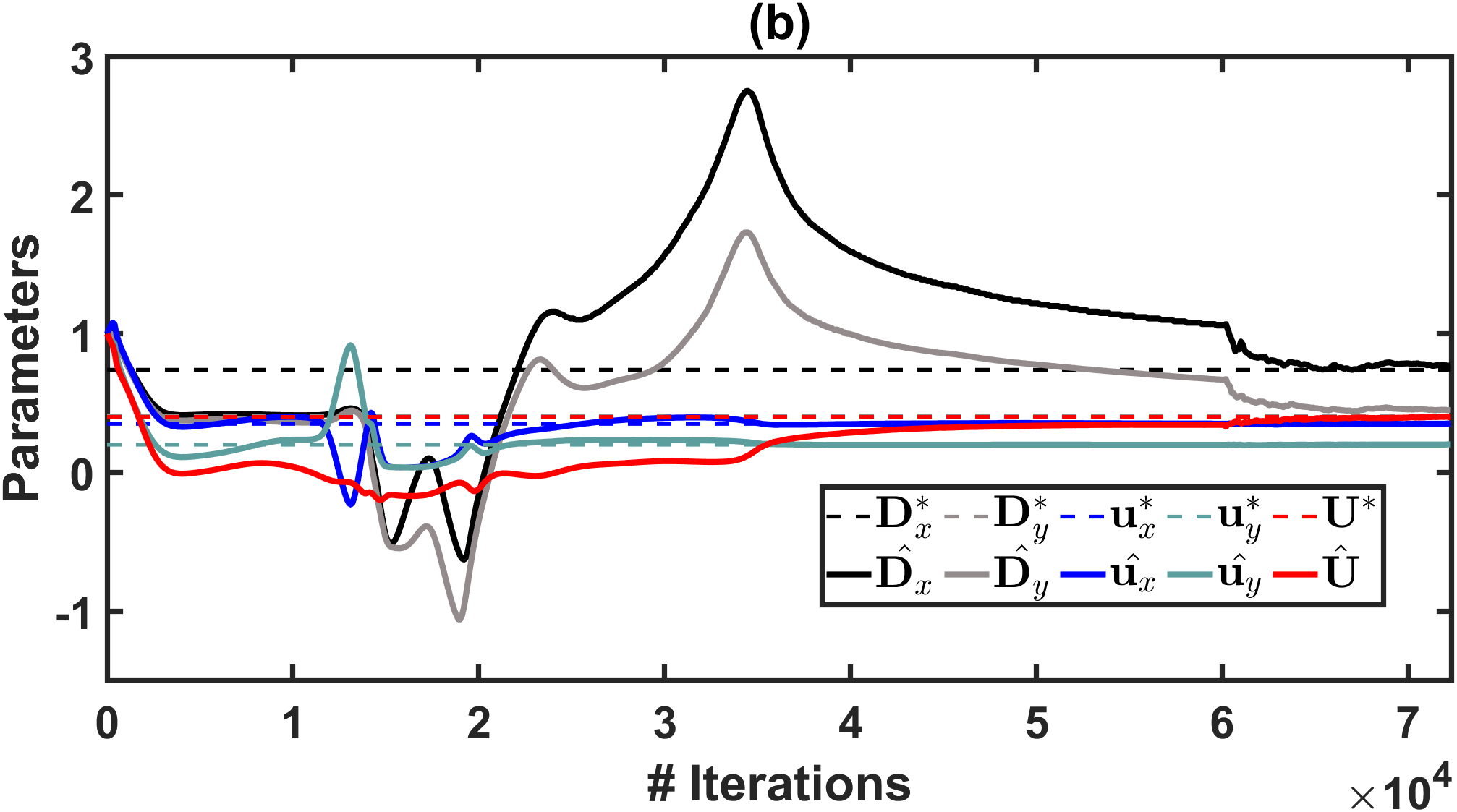}
\end{tabular}
\caption{(a) Dotted lines represent the constant parameter vector, $\bm{\theta}^{*} = [0.74\,\, 0.41\,\, 0.35\,\,0.2\,\, 0.4]^\T$, utilized for generating the training dataset for Case Study 3. Solid lines correspond to the temporally perturbed parameter vector, $\bm{\theta}(t) = [\textnormal{D}_x(t)\,\,  \textnormal{D}_y(t)\,\, \textnormal{u}_x(t)\,\, \textnormal{u}_y(t)\,\, \textnormal{U}(t)]^\T$,  employed for generating the training dataset for Case Study 4. (b) Parameter learning and convergence process. Predicted vector for the five model parameters, $\hat{\bm{\theta}} = [0.771\,\, 0.449\,\, 0.351\,\,0.2\,\, 0.4]^\T$. True vector used for generating the training dataset, $\bm{\theta}^{*} = [0.74\,\, 0.41\,\, 0.35\,\,0.2\,\, 0.4]^\T$.}
\label{fig:Noisy_2D_Data}
\end{figure}

To assess once more the performance of PiNNs across their entire range of functional capabilities, a brief example employing simulated noisy data is conducted. The ANN architecture and training properties align with the assumptions delineated in \Cref{sec:Case_3}, with the sole modification being the inclusion of training data points perturbed by some form of noise (as further elaborated in \Cref{sec:Case_2}). A model error equivalent to $\delta_{\varphi}=5\%$ and a correlation time of $\zeta_{\varphi}= 0.005$ are intentionally chosen to introduce temporal fluctuations of approximately 15$\%$ around the nominal values, $\bm{\theta}^{*} = [\textnormal{D}_x^{*}\,\, \textnormal{D}_y^{*}\,\, \textnormal{u}_x^{*}\,\, \textnormal{u}_y^{*}\,\, \textnormal{U}^{*}]^\T = [0.74\,\, 0.41\,\, 0.35\,\,0.2\,\, 0.4]^\T$, constituting a vector $\bm{\theta}(t) = [\textnormal{D}_x(t)\,\,  \textnormal{D}_y(t)\,\, \textnormal{u}_x(t)\,\, \textnormal{u}_y(t)\,\, \textnormal{U}(t)]^\T = [\textnormal{D}_x^{*} + \Delta \textnormal{D}_x(t)\,\, \textnormal{D}_y^{*} + \Delta \textnormal{D}_y(t)\,\, \textnormal{u}_x^{*} + \Delta \textnormal{u}_x(t)\,\, \textnormal{u}_y^{*} + \Delta \textnormal{u}_y(t)\,\, \textnormal{U}^{*} + \Delta \textnormal{U}(t)]^\T$. This variability is depicted in \Cref{fig:Noisy_2D_Data}(a), expressing temporal fluctuations around the average values throughout the simulation duration. The remarkable proficiency of PiNNs in learning the unspecified coefficients is evident in \Cref{fig:Noisy_2D_Data}(b), where the predicted parameter vector, $\hat{\bm{\theta}} = [\hat{\textnormal{D}}_x\,\, \hat{\textnormal{D}}_y\,\, \hat{\textnormal{u}}_x\,\, \hat{\textnormal{u}}_y\,\, \hat{\textnormal{U}}]^\T = [0.771\,\, 0.449\,\, 0.351\,\,0.2\,\, 0.4]^\T$, approximately matches $\bm{\theta}^{*}$. This validation, supported by the convergence of the model parameters to their nominal values, underscores the potential of PiNNs as a powerful tool for modeling complex physical systems with inherent noise and randomness. 

\subsubsection{Parameter learning for Troy Fire event in California}
\label{sec:Case_5}

\begin{figure}[tbp]
\centering
\includegraphics[width=.75\textwidth]{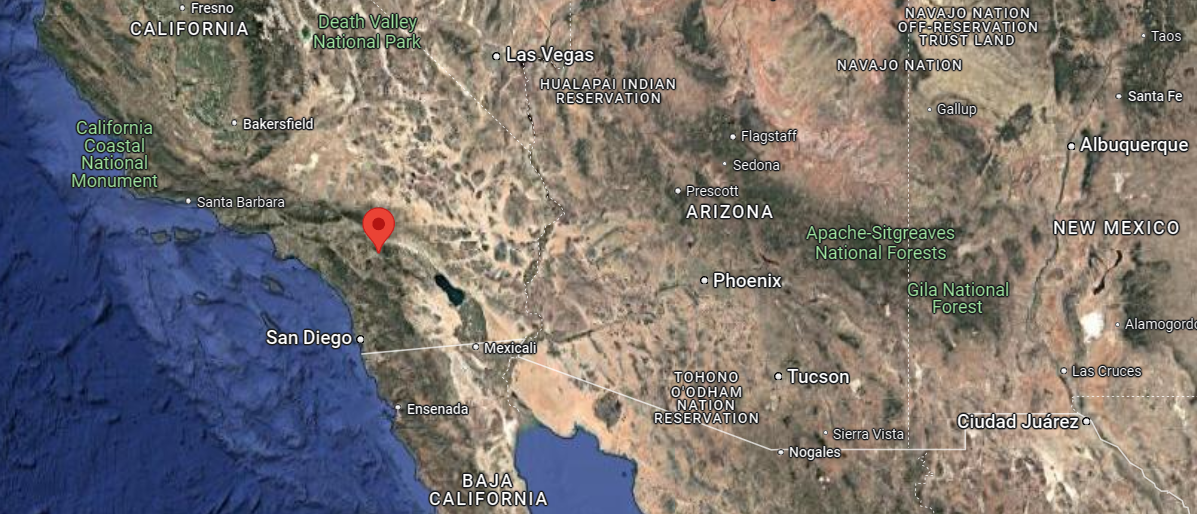}
\caption{Location of the Troy Fire event is indicated by the red symbol. [Source: Google Maps].}
\label{fig:Troy_location}
\end{figure}

After testing the functionality of PiNNs on four simulated scenarios leveraging synthetic data--both with and without noise--and verifying the effectiveness of the proposed approach in one- and two-dimensional case studies, the same method is then applied to a real-world fire scenario. One of the most notable wildland fires in recent history, known as the Troy Fire, occurred in California on June 19, 2002. This natural disaster burned 1.188 acres of land, starting in the Laguna Mountains of San Diego County, California (eastern edge of the Cleveland National Forest), as shown in \Cref{fig:Troy_location}. This historical forest fire is particularly representative, despite the limited data available on temperature, fuel, topography, and weather conditions (e.g., air temperature, relative humidity, wind speed, and wind direction), as it still offers valuable insights for analysis. 

\begin{figure}[tbp]
\centering
\includegraphics[width=.55\textwidth]{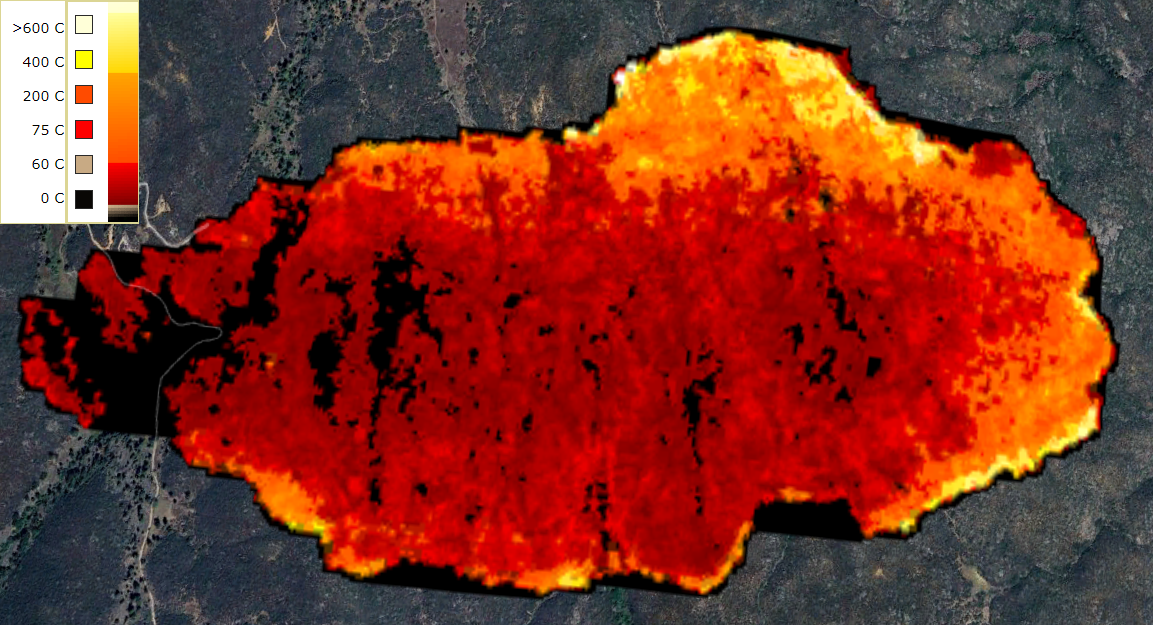}
\caption{Thermal image (in $^\circ\text{C}$
) of the Troy Fire event at 2:46 PM serves as the initial condition for our simulation \citep{USDA2019} [Source: Google Earth].}
\label{fig:Troy_initial}
\end{figure}

A key source of information is the thermal imagery, partially transmitted by the Pacific Southwest Research Station (PSW) Airborne Sciences Aircraft (Piper Navajo remote sensing aircraft) \citep{PacificSouthwest2007} via satellite communications in near real-time. The FireMapper\texttrademark \ system \citep{Riggan2003, RigganJ2003}, a thermal-imaging radiometer, detects thermal-infrared radiation emitted from the flame surface, effectively penetrating the smoke plume. As a result, 62 thermal images (24 of which are available) \citep{PacificSouthwest2007} were collected between 1:30 PM and 6:31 PM on the day of the fire, capturing accurate temperature distributions of the burning area. During the initial period of fire expansion (from 1:30 PM to 2:32 PM), the flames primarily spread in a southeast direction. Subsequently (from 2:39 PM to 6:31 PM), the combined effects of topography and fuel type caused a rapid acceleration of the fire toward the northeast. To collect reliable data for the neural network training phase, we analyzed the firefront's progression between two consecutive time points, from 2:46 PM to 2:52 PM (a 6-minute interval). This timeframe is appropriate, as our physics-based approach manipulates small-scale spatial and temporal resolutions compared to others, which range from hours to days and across thousand of kilometers \citep{Shadrin2024}. \Cref{fig:Troy_initial} depicts the thermal image of the Troy Fire at 2:46 PM, which is then applied as an initial condition for our simulation. Warmer colors indicate intense or active combustion, while darker colors signify lower temperature values and more heated bare ground. During this interval, the ambient conditions were relatively calm, and the topography was flatter.

\begin{figure}[tbp]
\centering
\includegraphics[width=.55\textwidth]{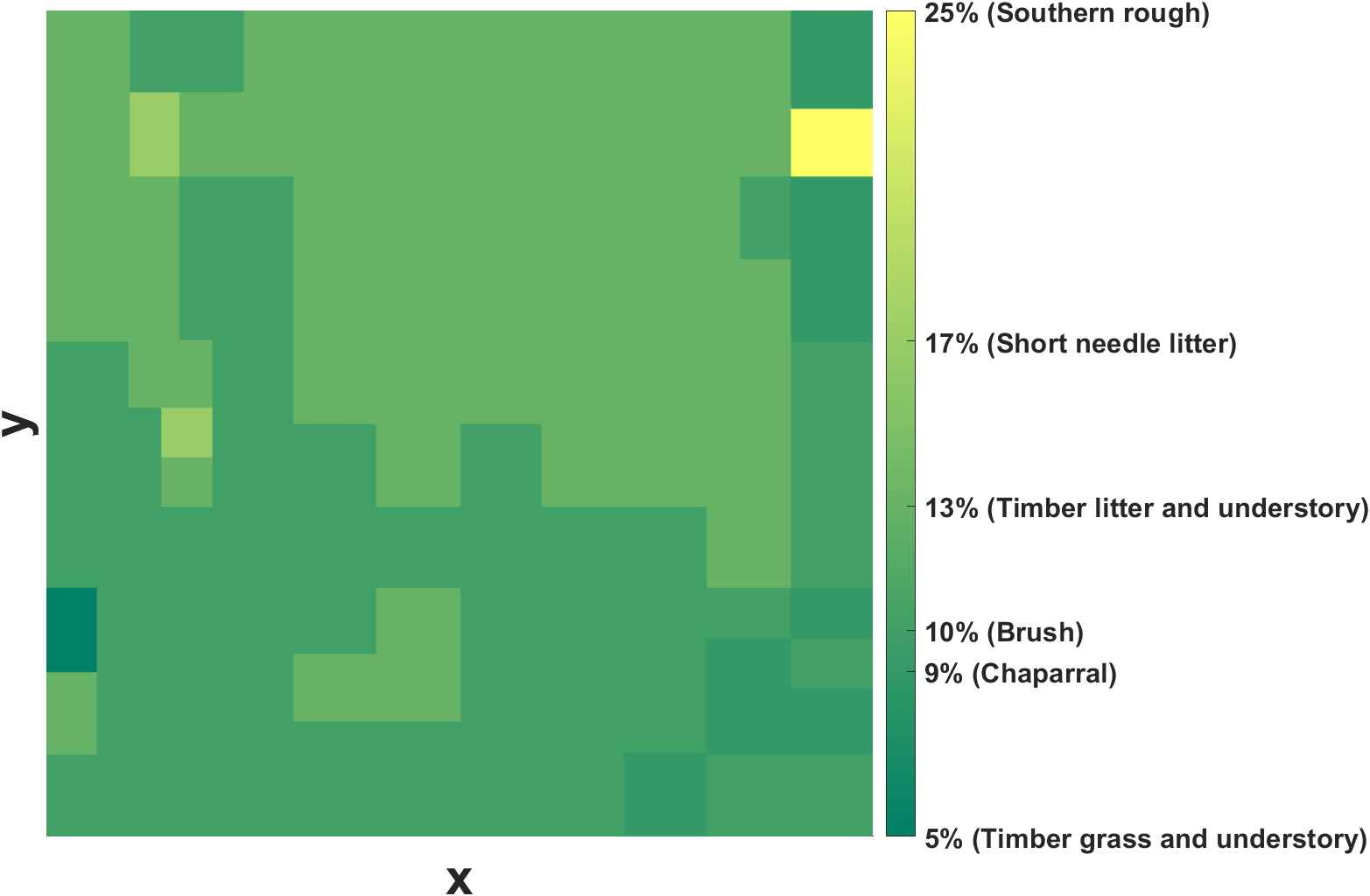}
\caption{Vegetation map showing moisture content ($\%$) at the initial time ($t =$ 0) of simulation \citep{Lo2012}.}
\label{fig:Troy_fuel}
\end{figure}

The distribution of fuel, along with various fuel types, geometries, and moisture content, is one of the most significant factors influencing wildfire behavior. The Troy forest region features diverse fuel categories, primarily consisting of short grass, timber grass and understory, chaparral, and brush. The US Forest Service's LANDFIRE website \citep{USDA2019} provides detailed raster data and vegetation maps based on Anderson’s 13 fuel model (13FBFM) \citep{Burgan1984, Rothermel1972, ScottBurgan2005}. This model classifies fuels into four general groups: grass (models 1 to 3), chaparral and shrub (models 4 to 7), timber litter (models 8 to 10), and logging slash (models 11 to 13). It offers detailed information regarding surface-to-volume ratios, fuel loading, fuel bed depth, and moisture of extinction. Each fuel type is associated with a specific color code and description corresponding to these attributes. However, since our simplified physics-based approach focuses only on the endothermic and exothermic phases of fuel across a uniform distribution \citep{Vogiatzoglou2024}, we will not incorporate the high-resolution LANDFIRE vegetation map. Instead, we generate a fuel map for our simulation based on a low-resolution image, where the moisture of extinction ($S_{1,0}$) is the most critical factor. \Cref{fig:Troy_fuel} illustrates the fuel moisture content distribution ($\%$) across the simulated domain at the initial time ($t =$ 0), and \Cref{table:Moisture_content} provides the corresponding values.

\begin{table}[tbp]
\centering
\caption{Moisture content for various fuel types (as classified by the 13FBFM \citep{ScottBurgan2005}) across the simulated domain.}
\label{table:Moisture_content}
\begin{tabular}{ccccccc}
\toprule
\textbf{Fuel type} & \makecell{\textbf{Timber litter} \\ \textbf{and understory}} & 
\textbf{Brush} &
\makecell{\textbf{Short needle} \\ \textbf{litter}} & \makecell{\textbf{Timber grass} \\ \textbf{and understory}} &
\textbf{Chaparral} & \textbf{Southern rough} \\
\midrule
\makecell{\textbf{Moisture content} \\ \textbf{$S_{1,0}$ ($\%$)}} & 13 & 10 & 17 & 5 & 9 & 25 \\
\bottomrule
\end{tabular}
\end{table}

\emph{Case Study 5} aims to replicate a real wildfire scenario, specifically the Troy Fire, following methodologies established in existing literature \citep{Alessandri2021, Heui2012, Lo2012}. Given the challenge of acquiring detailed spatiotemporal data--such as temperature and fuel measurements--an inverse ``approximation'' strategy was necessary. In the first stage, thermal images of the Troy Fire were processed using MATLAB, converting them into numerical matrices for integration into our simulations \citep{Heui2012}. We utilized thermal images captured at 2:46 PM (as the initial temperature condition, shown in \Cref{fig:Troy_initial}) and 2:52 PM (representing the final temperature distribution), along with the initial vegetation map (\Cref{fig:Troy_fuel}). These inputs were fed into a least squares minimization algorithm to compare the predicted and observed final temperature distributions across multiple iterations \citep{Heui2012}. Through this approach, we successfully identified the optimal values for the five uncertain model parameters, $\bm{\theta}^{*} = [\textnormal{D}_x^{*}\,\, \textnormal{D}_y^{*}\,\ \textnormal{u}_x^{*}\,\,\textnormal{u}_y^{*}\,\,\textnormal{U}^{*}]^\T$. These values not only matched the final observed temperature distribution but also accurately simulated the wildfire's progression, thereby demonstrating the potential of our method to approximate real-world fire behavior despite the absence of comprehensive field data.

Due to lack of spatiotemporal temperature and fuel measurements, an alternative approach to approximate wildfire dynamics was implemented. Real-world data is often sparse or unavailable, making direct training of PiNNs impractical. By converting thermal images from the Troy Fire into numerical matrices and by leveraging the optimal values from the optimization algorithm, we created a surrogate dataset that approximates the actual fire behavior. Once again, the applicability of the proposed PiNNs framework remains consistent in this case study, with the only difference being the incorporation of a surrogate training dataset derived from actual thermal images. Both \Cref{fig:PiNN_Diagram} and the loss terms are designed similarly to adhere to physical laws throughout the learning process. This approach enabled us to simulate the fire’s evolution using real initial and final conditions, ensuring that the identified parameters closely matched the actual event, despite the lack of detailed measurements.

In this context, the primary objective of this practical case study is to identify the five unknown model parameters, leveraging simplified physical, topographical, and fuel assumptions, along with a small-scale PiNN architecture. Consequently, the ANN structure remains consistent with that employed in \emph{Case Study 3} (see \Cref{sec:Case_3}) and \emph{Case Study 4} (see \Cref{sec:Case_4}), with the notation set as follows: $L$ = 5, $N_{0}$ = 3, $N_{\ell}$ = 20 for $\ell =$ 1, 2, 3, 4, and
$N_{5}$ = 3. A logistic sigmoid AF is applied under the Adam algorithm with a learning rate of 0.0003 for 60.000 iterations, followed by the L-BFGS optimizer. The computational domain is governed by open outflow boundary conditions, ensuring that the firefront can propagate smoothly beyond the domain's boundaries without causing significant backward influence \citep{Papanastasiou1992, Karniadakis2014}. The initial conditions are derived directly from the thermal image and vegetation coverage, keeping the entire model consistent within a non-dimensional framework. Throughout the learning algorithm, the model equations were included to provide constrained estimates that satisfy the physical components of the modeling approach.

\begin{figure}[tbp]
\centering
\begin{tabular}{c} \includegraphics[width=.55\textwidth]{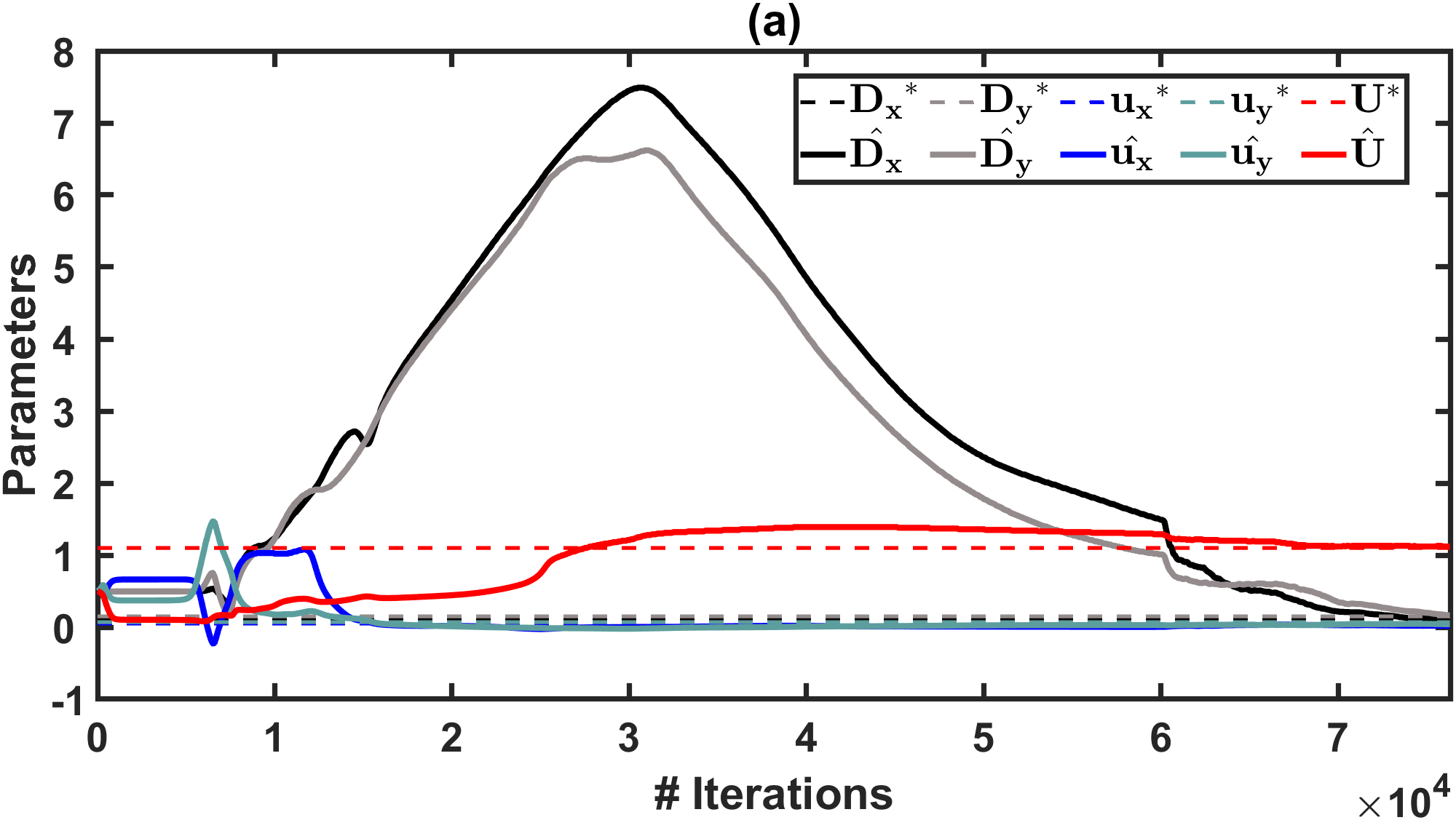} \\
\includegraphics[width=0.4\textwidth]{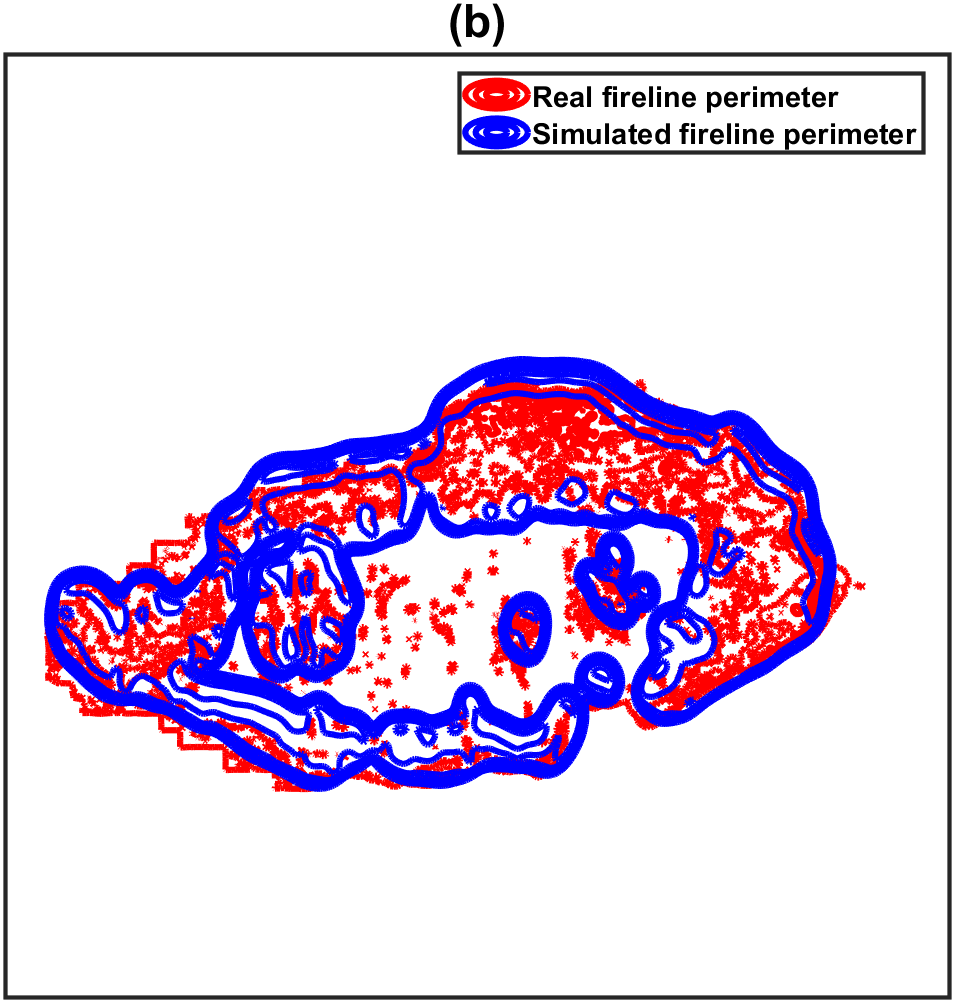} 
\end{tabular}
\caption{(a) Parameter learning and convergence process. Predicted vector for the five model parameters, $\hat{\bm{\theta}} = [0.102\,\, 0.165\,\, 0.022\,\,0.054\,\, 1.12]^\T$. True vector used for generating the training dataset, $\bm{\theta}^{*} = [0.1\,\, 0.15\,\, 0.05\,\,0.07\,\, 1.1]^\T$. (b) Comparison between the real (Troy Fire) and simulated fireline perimeter at the final time instant ($t =$ 1).}
\label{fig:Troy_convergence}
\end{figure}

The surrogate training dataset was generated using the extracted optimal values for the model parameters, denoted as $\bm{\theta}^{*} = [\textnormal{D}_x^{*}\,\, \textnormal{D}_y^{*}\,\, \textnormal{u}_x^{*}\,\, \textnormal{u}_y^{*}\,\, \textnormal{U}^{*}]^\T = \left[0.1\,\, 0.15\,\, 0.05\,\, 0.07\,\, 1.1\right]^\T$. This dataset contains 27.716 spatiotemporal data points corresponding to temperature and fuel measurements--fewer than in the previous simulated cases--yet suitable for the lightweight ANN architecture employed. Starting with initial estimates for all parameters set to 0.5, the network efficiently converged to the optimal values, $\bm{\theta}^{*}$. As shown in \Cref{fig:Troy_convergence}(a), all parameters being identified nearly converged to their optimal values, since $\hat{\bm{\theta}} = [\hat{\textnormal{D}}_x\,\, \hat{\textnormal{D}}_y\,\, \hat{\textnormal{u}}_x\,\, \hat{\textnormal{u}}_y\,\, \hat{\textnormal{U}}]^\T = [0.102\,\, 0.165\,\, 0.022\,\,0.054\,\, 1.12]^\T \approx \bm{\theta}^{*}$, during the course of iterations (9 h of CPU
time). Notably, the advection and overall heat transfer coefficients ($\textnormal{u}_x, \textnormal{u}_y$, and $\textnormal{U}$) converged within the first thousand iterations, while the dispersion coefficients ($\textnormal{D}_x, \textnormal{D}_y$) required additional iterations taking advantage of the L-BFGS optimizer to achieve higher precision. Although the wind speed components slightly deviate from the optimal values, this is acceptable given the lightweight neural network architecture and the extremely noisy input data. The accuracy of the parameter identification process is further demonstrated in \Cref{fig:Troy_convergence}(b), which compares the simulated and actual (obtained from the thermal images) fireline perimeters at the final time step ($t =$ 1) of the simulation. The close agreement between these perimeters illustrates the effectiveness of this method in managing real-world wildfire events and uncovering the underlying physical parameters.

The parameter learning process using neural networks offers a powerful framework for identifying model parameters, even in the case of an actual fire event. However, several challenges can cause deviations from the expected optimal results. Specifically, empirical data, such as thermal images, often contain significant noise due to the limitations of the equipment used for data collection, which introduces inaccuracies during the training phase. As a result, the model may struggle to accurately capture the true underlying patterns, leading to prediction errors. Furthermore, both the training datasets and the neural network architecture may be too simplistic to effectively scale to larger spatial and temporal events. These factors highlight the need for more sophisticated models and higher-quality data to improve the accuracy and scalability of predictions.

\section{Discussion  and conclusions}
\label{sec:Discussion_and_Conclusions}

Wildfire-related natural disasters are among the most intricate environmental phenomena to model, as they initiate complex physical processes while being subject to dynamically evolving atmospheric conditions. The modeling formulation and ambient variations introduce numerous uncertainties, enhancing the risk of inadequate predictions of wildfire spread. This paper highlights the importance of PiNNs (with perspectives on data-driven modeling) for estimating critical unknown parameters that are often difficult to measure in real-world scenarios. The innovative aspect lies in restricting the learning process to obey the physical constraints of the dynamic system directly into the
training phase, thereby capturing the physics underlying it. In a general sense, this framework can serve as an offline tool (proactive strategy) for simulating upcoming wildfire events across various scenarios and by producing multiple risk maps.

All the examined case studies showcase the utility of PiNNs in uncovering the unknown model parameters across a wide spectrum of representative examples, encompassing both one- and two-dimensional firefront propagation events. Particularly, PiNNs unveiled great efficiency in identifying nominal values using both synthetic and empirical wildfire event data from the Troy fire in California. Despite the physics-consistent nature of the model, the simplicity of the ANN architecture aligned well with the size of the training dataset. Moreover, this revolutionary approach proved to be functional even when dealing with simulated data and highly noisy temperature measurements obtained from thermal images. The findings underline the versatility and operability of PiNNs in addressing complex modeling tasks, boosting their performance as both inverse and forward predictors.

Although PiNNs can effectively learn the required wildfire spreading parameters, 
there are limitations related to the complexity of the training phase. As a result, small-scale spatial and temporal domains were chosen to adequately demonstrate the core physical concepts while minimizing computational demands. In this context, Hierarchical Bayesian learning \citep{Sedehi2024} will lay the groundwork for quantifying inherent uncertainties across these consecutive spatial and temporal intervals, accounting for the general uncertainty arising from modeling assumptions and dynamically evolving environmental conditions. Measurements often represent averaged values over specific regions and may not provide direct insights for the spatiotemporal description of all network output quantities of interest. Mixed neural networks, consisting of a convolutional neural network (CNN) for processing satellite imagery alongside a PiNN, portray a notable leap in the direction of improving the broader framework for wildfire management within the field of artificial intelligence and machine learning.





\bibliographystyle{elsarticle-num}
\bibliography{PINN_Wildfire_Spreading}



\end{document}